\documentclass[final,5p,times,twocolumn,compress]{elsarticle}

\usepackage{graphicx}
\usepackage{lineno}
\usepackage{amsthm,amsmath}
\usepackage{mathrsfs}
\usepackage[utf8]{inputenc}
\usepackage{algorithm}
\usepackage{algpseudocode}
\usepackage{hyperref}
\usepackage{epstopdf}
\usepackage{multirow}
\usepackage{booktabs}
\usepackage{xcolor}
\usepackage{xpatch}
\usepackage{subcaption}

\makeatletter
\def\changeBibColor#1{%
  \in@{#1}{ }%  list of colored bib items
  \ifin@\normalcolor\else\color{blue}\fi
}

\xpatchcmd\@bibitem
  {\item}
  {\changeBibColor{#1}\item}
  {}{\fail}

\xpatchcmd\@lbibitem
  {\item}
  {\changeBibColor{#2}\item}
  {}{\fail}
\makeatother

\biboptions{square,numbers,sort}

\journal{Elsevier}

\begin{document}
\begin{frontmatter}
\title{Feature Transformation Enhanced Jacobi Polynomial Graph Filtering for Graph Anomaly Detection}

\cortext[cor1]{Corresponding author at: Institute of Artificial Intelligence, Fujian University of Technology, Fuzhou, China.}
\author[org2]{Xiang Wang }
\author[org1]{Zhijun Cheng}
\author[org1]{Zhenyu Meng \corref{cor1}}
\ead{mzy1314@gmail.com}
%%\address[author1]{School of Computer Science and Mathematics, Fujian University of Technology, Fuzhou, China}

\affiliation[org1]{organization={Institute of Artificial Intelligence, Fujian University of Technology}, city={Fuzhou}, country={China}}%%Department and Organization
\affiliation[org2]{organization={School of Artificial Intelligence and Transportation Engineering, Fujian University of Technology}, city={Fuzhou}, country={China}}

% Here goes the abstract
\begin{abstract}
In recent years, graph anomaly detection (GAD) based on frequency-domain filtering have achieved promising results. However, existing approaches still face three major challenges: First, they use static basic function to constructed graph filter which cannot effectively adapt to the frequency-domain distribution of graph data. Second, they fail to adequately consider the importance information of each attribute in the node feature vector, leading to the loss of fine-grained information. Third, they insufficiently utilize node labels for GAD. To address these issues, this paper proposes a novel graph anomaly detection method called JPGFN (Feature Transformation Enhanced Jacobi Polynomial Graph Filtering Network). First, a Feature Separation Transformation Network (FSTNN) is developed to better learn fine-grained node features by feature separation and applying nonlinear transformations to node features across different dimensions. Second, an adaptive Jacobi polynomial graph filtering module is constructed based on Jacobi polynomials to adaptively capture complex frequency-domain features of graph signals. Finally, a node label constraint module is developed to facilitate the use of node labels and enhance the performance of GAD. Experimental results on multiple real-world datasets demonstrate that the proposed method significantly outperforms mainstream approaches.
\end{abstract}

%\nocite{*}


\end{frontmatter}

% Main text
\section{Introduction}\label{1}
With the rapid development of the digital society, graph data has found widespread application across multiple fields. For example, in social networks, graph data can be used to analyze the complex interactions between different users \cite{LIU2026112935}. In the financial sector, graph data can effectively represent user transactions and capital flows \cite{2motie2024}. And in bioinformatics, protein interaction networks and regulatory relationships between genes can all be represented using graph data \cite{3cui2025}. In these applications, the occurrence of various anomalous events has given rise to the critical task of graph anomaly detection, such as fake accounts in social networks, fraudulent activities in financial transactions, and protein abnormalities in biological networks. These phenomena may pose potential risks to users or living organisms, making it imperative to design effective graph anomaly detection methods to identify such events \cite{45yuan2025comprehensive}.

To detect anomalous events and objects, conventional deep learning approaches typically express objects as attribute vectors and subsequently perform anomaly detection in the feature space \cite{4Huang2025,5HojjatiHA2024}. However, these methods often fail to account for the complex interactive relationships between different entities, resulting in poor detection performance \cite{HEVAPATHIGE2026108869}. With the rapid development of graph neural networks, anomaly detection methods based on graph neural networks have made significant progress \cite{49li2017radar}. Compared to traditional methods, the effectiveness of graph neural networks lies in their feature aggregation mechanism, which can effectively capture both node features and topological information, thereby more effectively identifying anomalous nodes in complex networks \cite{20KipfW2017GCN,22Velickovic2018GAT,7WangDDM2025}. The powerful graph representation learning capabilities enabled by this mechanism allow graph neural networks to demonstrate superior performance over traditional deep learning methods in graph anomaly detection tasks, making them the mainstream approach in this field today \cite{50ma2021comprehensive}.

Current mainstream GAD methods based on graph neural networks are primarily classified into two categories: spatial-domain and frequency-domain approaches \cite{50ma2021comprehensive}. The former method mainly rely on message passing mechanisms to learn graph features by iteratively aggregating neighborhood information \cite{10gao2023GDN,9MesgaranH2024GFCN,WANG2025114039}. Although these methods are effective at learning the features and structural information of graph data, they neglect the learning of frequency-domain features in graph signals, resulting in poor performance in detecting anomalous nodes \cite{11zhang2025}. In contrast, spectral domain-based methods, which can capture complex frequency domain information, often demonstrate superior detection performance \cite{12ding2025novel}. These methods primarily employ polynomial approximation strategies to design graph filters capable of effectively learning frequency-domain features. Mainstream approaches include CHRN \cite{41gao2023addressing}, SEC-GFD \cite{14XuWWWZW24SECGFD} and DSGAD \cite{15zheng2025DSGAD} etc.

However, spectral domain-based methods still have limitations. First, these methods employ static graph filters, which are unable to effectively adapt to complex spectral distributions of real-world graph data.  For example, Figure \ref{fig1} shows two graphs containing anomalous nodes and their corresponding spectral distributions, revealing significant differences in spectral distributions across different graphs. Existing spectral domain-based methods employ static basis function to construct graph filters for GAD, lacking flexibility and making it difficult to fully capture frequency-domain information. Second,existing methods do not sufficiently consider the varying importance of different node features in GAD tasks. In Figure \ref{fig1}, node features are represented as vectors where features of different dimension have varying important in GAD tasks. Taking the real-world Elliptic dataset, a Bitcoin transaction network where nodes represent users as an example, node features include transaction frequency, age, and address etc. Anomalous nodes typically exhibit multiple consecutive transactions within a short timeframe, strongly correlating with ``transaction frequency'' while showing weaker associations with features like ``age'' or ``address''. Existing methods often neglect the impact of such factor, leading to the loss of fine-grained node features. Third, these methods often fail to effectively utilize node labels, which can provide additional information for GAD.

% \begin{figure}[!htbp]
% \centering
% \includegraphics[width=0.5\textwidth]{figs/fig1.png}
% \caption{Graph data and its corresponding frequency domain distribution.}
% \label{fig1}
% \end{figure}

Therefore, this paper proposes a GAD method, namely JPGFN (Feature Transformation Enhanced Jacobi Polynomial Graph Filtering Network). JPGFN primarily consists of FSTNN, adaptive Jacobi polynomial graph filtering module, and node label constraint module. The FSTNN module uses feature separation and nonlinear transformation strategy to learn fine-grained features. The adaptive Jacobi polynomial graph filtering module conducts graph filtering based on parameterized Jacobi basis, enabling flexible adaptation to the frequency domain distribution of graph data and thus extracting richer information. The node label constraint module enhances the model's anomaly detection performance by comparing the consistency between the representations of a central node and its surrounding nodes. The primary highlights of this paper are summarized as follows:
\begin{figure}
	\centering
	\includegraphics[width=.99\columnwidth]{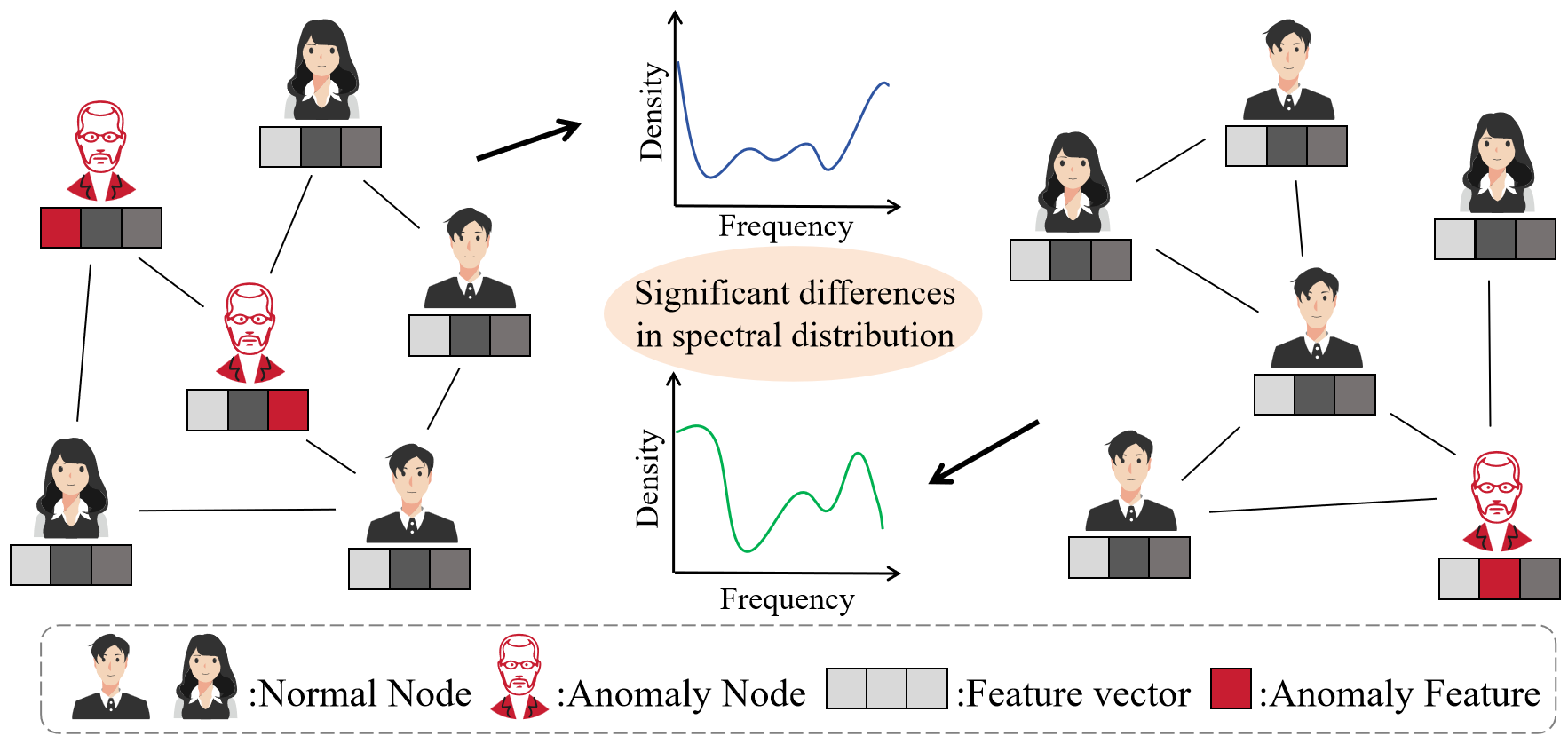}
	\caption{Graph data and its corresponding frequency domain distribution.}
	\label{fig1}
\end{figure}

\begin{itemize}
    \item To better learn fine-grained features, a Feature Separation Transformation Network (FSTNN) is developed.
    \item To more effectively learn complex frequency domain information of graph signals, an adaptive graph filtering module based on Jacobi polynomial is designed.
    \item A node label constraint module is constructed to sufficiently use node labels for GAD.
    \item Experimental results across multiple real-world datasets demonstrate that the proposed method significantly outperforms mainstream approaches.
\end{itemize}

\section{Related Work}
\subsection{Graph Neural Networks}
The rapid development of graph neural networks (GNNs) has addressed the challenge faced by traditional deep learning methods in effectively modeling the complex interactions between different entities. Current GNNs approaches have realized significant outcomes in various tasks, including node classification \cite{16eliasof2024global,17ratna2025inclusive} and graph classification \cite{18wang2024graph,19qian2025exploring}. Among these, GCN \cite{20KipfW2017GCN} stand as a representative approach, learning node representations by aggregating first-order neighbor features. Subsequently, numerous variants emerged based on the message-passing paradigm, including GraphSAGE \cite{21hamilton2017GraphSAGE}, GAT \cite{22Velickovic2018GAT}, GIN \cite{23xu2018GIN}, and PMP \cite{24he2024polarized}. Concurrently, some researchers perform graph data analysis from the perspective of spectral domain filtering, leading to methods such as ChebyNet \cite{25defferrard2016}, SEA-GWNN \cite{27deb2024sea}, NFGNN \cite{28ZhengZLLZ24NFGNN} and AGFNN \cite{26zhang2024beyond}.

Although existing GNNs methods have achieved remarkable performance across a wide range of graph data mining tasks, their effectiveness generally relies on the assumption that neighboring nodes exhibit similar class labels and features. However, this assumption often does not hold in the context of graph anomaly detection, as anomalous nodes tend to interact with normal nodes rather than with nodes sharing similar characteristics. Therefore, directly applying general-purpose GNNs to graph anomaly detection tasks often yields suboptimal performance. It is necessary to develop dedicated graph anomaly detection methods to enhance the effectiveness and accuracy of anomaly detection.

\subsection{Graph Anomaly Detection}
Recently, GNNs-based methods for GAD have gained widespread attention. They are typically classified into two categories: spatial-domain and frequency-domain approaches.

The GAD approaches based on spatial-domain obtain node embeddings through iterative aggregation of information from neighboring nodes. For example, PC-GNN addresses the problem of category imbalance by designing subgraph sampling and neighbor sampling strategies \cite{37liu2021}. GFCN \cite{9MesgaranH2024GFCN} introduces skip connections to captures long-range node features. DiG-In-GNN \cite{29zhang2024dig} incorporates a feature guidance module and a neighbor selection module to address feature inconsistency and structural inconsistency issues. CIE-GAD \cite{30huang2025correlation} employs hypergraph transformation to uncover association patterns among nodes. HEAug \cite{11278786} overcomes the issue of high-category homophily variance (CHV) where benign nodes are highly homophilic, but anomalies are not—by generating low-CHV links while using original edges as an auxiliary. Although these methods leverage node attribute information and structures to effectively identify anomalies, they cannot learn the frequency-domain information of graph data.

The spectral-domain methods use graph filtering operations to extract frequency-domain information from graph data. For example, BWGNN \cite{13tang2022} introduces wavelet function family to construct filters that addresses the spectral ``right shift'' phenomenon caused by anomalous nodes. SEC-GFD \cite{14XuWWWZW24SECGFD} proposes a hybrid filtering module based on Beta wavelet functions to address heterophily issues. AHFAN \cite{7WangDDM2025} designed a graph filtering module based on Chebyshev polynomials and a node representation module based on attention mechanisms to addressing category inconsistency and semantic inconsistency issues, respectively. DSGAD \cite{15zheng2025DSGAD} addresses the issue of incomplete capture of anomalous information by traditional wavelet filters through the design of dynamic wavelet filters and dynamic fusion mechanism. EGNN \cite{liu2026modeling} develops a graph learning model based on energy-aware mechanism which can learn spectral characteristics via energy-driven feature aggregation. McGAD \cite{HUANG2026104338} employs two types of augmentation method based on structural consistency and learnable unsupervised consistency for GAD. Although these methods achieve satisfactory performance in GAD, they all employ static graph filters that struggle to adapt to the complex frequency-domain distributions of graph data. Furthermore, these approaches do not sufficiently consider the influence of different node features on anomaly detection. Additionally, they do not sufficiently leverage node label information.

\section{Preliminaries}
\subsection{Notations and Problem Definition}
An attributed graph is represented as $G = (V, X, A)$, where $\mathcal{V} = \{v_1, v_2, ..., v_n\}$ denotes the set of nodes, $\mathit{N} $ denotes the number of nodes.$\mathit{X} \in \mathit{R} ^{\mathit{N}\times \mathit{d}}$ denotes the feature matrix for graph nodes. $\mathit{d}$ is the feature dimension. $\mathit{A} \in \mathit{R}^{\mathit{N}\times \mathit{N}}$ indicates the adjacency matrix for attributed graph. $\mathit{A} _{\mathit{i}, \mathit{j}} = 1$ indicates that there is an edge connecting nodes $\mathit{v} _{\mathit{i} }$  to node $\mathit{v} _{\mathit{j} }$, otherwise, $\mathit{A}_{\mathit{i}, \mathit{j}} =0 $.

For a given attributed graph $\mathit{G}$ , every node is associated with a corresponding label $\mathit{y}=\left \{ 0,1 \right \}$ . Label 0 denotes normal nodes, while Label 1 indicates abnormal nodes, with significant differences observed between the features of these two node categories. This study focuses on semi-supervised anomaly detection on attributed graphs, specifically training a classifier to determine node abnormality given partial node labels.

\subsection{Graph Filtering}
Given a graph, its normalized Laplacian matrix is denoted as $\mathit{L} =\mathit{I}-\mathit{D}^{-1/2}  \mathit{A} \mathit{D} ^{-1/2}$ , $\mathit{D}$ is degree matrix of $\mathit{A}$ and $\mathit{I}$ is identity matrix. The matrix $\mathit{L}$ undergoes an eigenvalue decomposition to yield $\mathit{L}=\mathit{U}^{\mathit{T} } \mathit{\Lambda } \mathit{U^{T}}$, where $\mathit{\Lambda } =diag\left ( \left [ \lambda _{1},...,\lambda _{\mathit{N} } \right ]  \right )$ is the diagonal matrix composed of eigenvalues, and $\mathit{U} =\left [ \mathit{u}_{1},..., \mathit{u}_{\mathit{N} }   \right ]$  is the orthogonal matrix formed by the corresponding eigenvectors. Let $\mathit{U}^{T} \mathit{X}$ and $\mathit{U} \mathit{X}$ denote the Fourier transform of signal $\mathit{X}$ and its inverse transform, respectively. Graph filtering operations are defined by the graph filter $g \left (  \mathit{\Lambda}  \right )$. Graph filtering of signal $\mathit{X}$ is written as $\mathit{U} g \left (  \mathit{\Lambda}  \right )  \mathit{U}^{T} X$, where $g \left (  \mathit{\Lambda}  \right ) =diag\left ( g\left ( \lambda_{1} \right ), g\left ( \lambda_{2} \right )..., g\left ( \lambda_{N} \right ) \right )$. To avoid the high computational cost associated with eigenvalue decomposition, the filter function  $g \left (  \cdot   \right )$ is typically set as a polynomial with $K$-order, $g\left (  \mathit{\Lambda }    \right ) =\sum_{k=0}^{K} \theta _{k} \mathit{\Lambda }  ^{k}$, thereby simplifying the filtering operation to the following:
\begin{equation}
g\left ( \mathit{\Lambda }  \right )X=U\left ( \sum_{k=0}^{K} \theta _{k} \Lambda  ^{k} \right )U^{T}X = \sum_{k=0}^{K} \theta _{k}\mathit{L^{k} X}
\end{equation}

\section{Methodology}
\begin{figure*}[htbp]
\centering
\includegraphics[width=0.98\textwidth]{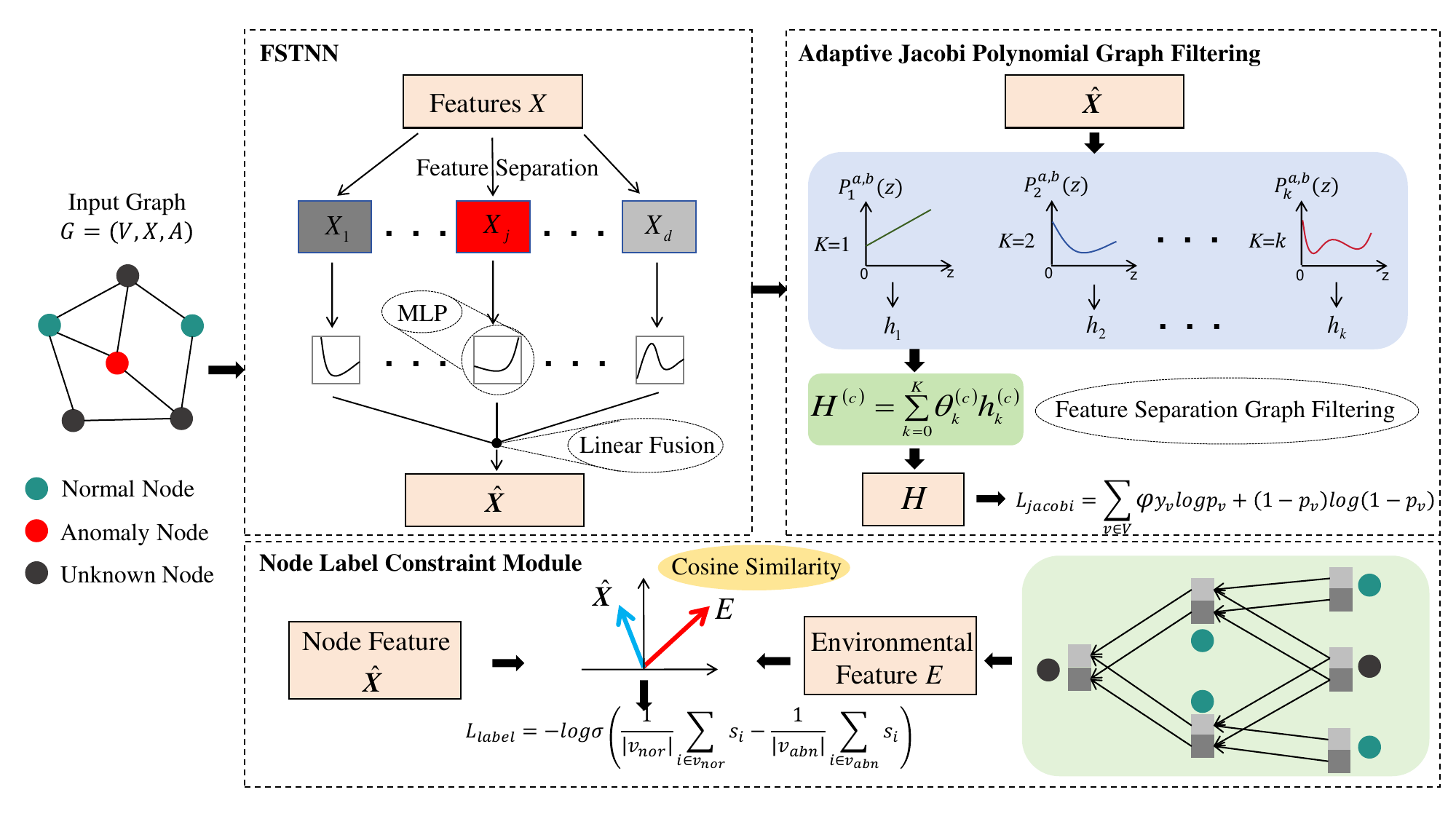}
\caption{Architecture diagram of JPGFN model.}
\label{fig2}
\end{figure*}
This section details the implementation of JPGFN which consists of three modules: Feature Separation Transformation Network (FSTNN), Adaptive Jacobi Polynomial Graph Filtering module, and Node Label Constraint Module. The model diagram is shown in Figure~\ref{fig2}.

\subsection{FSTNN}
Given that different node features impact anomaly detection differently, existing methods typically treat all features as equally important when feeding them into graph filters for frequency-domain information learning. This approach fails to effectively leverage node features for anomaly detection, thereby limiting the model's performance. To address this, nonlinear transformations must be applied to different node features prior to graph filtering to better learn fine-grained feature representations. For node $v_{i}$, the transformation is as follows:
\begin{equation}
X_{i,j}^{l+1}   =\phi _{j} \left (  X_{i,j}^{l}\right )
\end{equation}
where, $X_{i,j}^{l+1} $ is the transformed output of the $j$-th attribute for node $v_{i}$ at layer $l$ + 1, $\phi _{j} \left (\right )$ denotes the transformation operation for the $j$-th attribute.

Through feature separation transformation, the model can adaptively learn how different features influence anomaly detection, thereby better capturing fine-grained information. To enhance nonlinear transformation capabilities, the Universal Approximation Theorem establishes that a two-layer Multi-Layer Perceptron (MLP) can represents any function \cite{31lewicki2003approximation}. Therefore, an MLP is employed as the nonlinear transformation function to process node features. Specifically:
\begin{equation}
\phi _{j} \left (  X_{i,j}^{l}\right ) =\mathrm{MLP} \left (  X_{i,j}^{l} \right )
\end{equation}

Through feature separation transformation and learning process, the model can better extract anomaly-related feature information. After transformation, the $j$-th feature of node $v_{i}$ is represented as  $h_{i,j} $ . All transformed features of different dimensions are concatenated:
\begin{equation}
h_{i} =concat\left [ h_{i,1};h_{i,2};...;h_{i,d} \right ]
\end{equation}
here $concat\left (  \right )$  denotes concatenation, which fuses features across all dimensions through concatenation operations.After feature concatenation, the feature matrix of the entire graph data is $h$. Building upon this, a linear transformation is applied to further capture the semantic information of node features, yielding new node representations:
\begin{equation}
\hat{X} =Wh+b_h
\end{equation}
Here, $W$ and $b_h$ denote learnable matrix and bias term, respectively.

Through transformation and learning, this approach provides more discriminative node embeddings for subsequent graph filtering, enabling the model to better capture underlying anomaly patterns.

\subsection{Adaptive Jacobi Polynomial Graph Filtering}
\begin{figure}
	\centering
	\includegraphics[width=1.0\columnwidth]{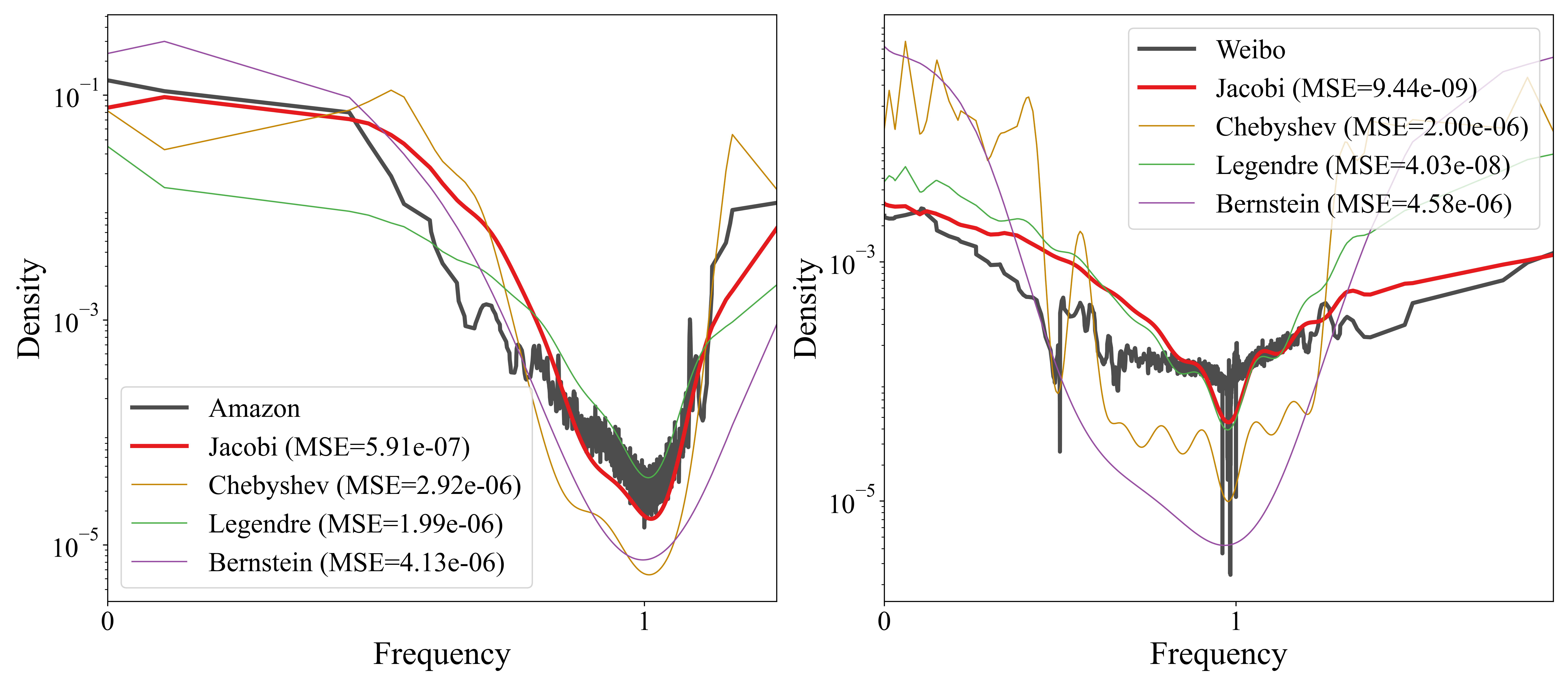}
	\caption{Signal density functions of different datasets and frequency-domain fitting results for different polynomial basis functions.}
	\label{fig3}
\end{figure}

For real-world graphs, their frequency domain distributions vary significantly. Existing spectral domain methods employ static basis functions for graph filtering, such as Chebyshev polynomials and Bernstein polynomials, which suffer from insufficient flexibility and struggle to adaptively learn different frequency distributions. Figure~\ref{fig3} illustrates the frequency domain distribution fitting results for different datasets using graph filters constructed with different polynomials. The MSE metric represents the mean squared error relative to the true graph data distribution. It is evident that the frequency domain curve learned by the Jacobi polynomial more close to the ground truth, with MSE significantly lower than that of other polynomial graph filters. This demonstrates that the Jacobi polynomial possesses superior frequency domain information learning capabilities. Therefore, this paper introduces graph filters constructed using the Jacobi polynomial basis, enabling flexible capture of frequency domain features for different graphs and more effective exploration of anomaly patterns in the frequency domain.

\subsubsection{Jacobi Polynomial Basis Functions}
The Jacobi polynomials are orthonormal polynomial family defined on [-1, 1] with weight function $\left ( 1-\lambda  \right )^{a}  \left ( 1+\lambda  \right ) ^{b}$, expressed as ($k\ge 2$):
\begin{equation}
P_{k}^{a,b} \left ( x \right ) = \left ( \alpha _{k}x+\beta _{k} \right ) P_{k-1}^{a,b} \left ( x \right )-\gamma _{k}P_{k-2}^{a,b} \left ( x \right )
\end{equation}
And when $k$ equal to 0 and 1, the Jacobi polynomials is:
\begin{equation}
\begin{split}
P_{0}^{a,b} \left ( x \right ) &=1\\
P_{1}^{a,b} \left ( x \right ) &=\frac{a-b}{2} +\frac{a+b+2}{2} x
\end{split}
\end{equation}
where
\begin{equation}
\begin{split}
\alpha _{k} &=\frac{\left ( 2k+a+b \right )\left ( 2k+a+b-1 \right )  }{2k\left ( k+a+b \right ) } \\
\beta _{k}  &=\frac{\left ( 2k+a+b-1 \right )\left ( a^{2}-b^{2}   \right )  }{2k\left ( k+a+b \right ) \left ( 2k+a+b-2 \right ) }\\
\gamma _{k}   &=\frac{\left ( k+a-1 \right )\left ( k+b-1 \right ) \left ( 2k+a+b \right )  }{k\left ( k+a+b \right ) \left ( 2k+a+b-2 \right ) }
\end{split}
\end{equation}
here, the range of values for $a$ and $b$ is $a,b> -1$. In fact, Jacobi polynomial is a more general class of polynomial family. This is due to the fact that by adjusting the parameters $a$ and $b$, Jacobi polynomial can degenerate into other polynomial families, such as Chebyshev polynomial, Legendre polynomial, and so on. Therefore, Compared to static basis functions, using Jacobi polynomials to construct graph filters offers greater flexibility, facilitating the capture of diverse frequency-domain information for different datasets. Exploiting the flexibility of Jacobi polynomials, we treat the parameters $a$ and $b$ as learnable variables, allowing the model to adaptively learn the complex spectral distributions of diverse graph datasets and thereby enhance its capability to model frequency-domain patterns.

\subsubsection{Graph Filter Construction}
Graph filters constructed based on Jacobi polynomials can capture frequency-domain information of various complex graph data. The graph filter constructed based on Jacobi polynomials is:
\begin{equation}
g_{k} \left ( \mathit{\Lambda} \right ) =\theta _{k} P_{k}^{a,b} \left ( 1-\mathit{\Lambda }  \right )
\end{equation}
here, $\mathit{\Lambda } =diag\left ( \left [ \lambda _{1},..., \lambda _{N} \right ]  \right )$ is the diagonal matrix formed by the eigenvalues of the Laplacian matrix $A$, $\theta _{k}$ represents the learnable polynomial coefficients, and $k$ denotes the order of the graph filter. Since the Jacobi polynomials are defined on [-1, 1], the function $g_{k} \left ( \mathit{\Lambda }  \right )$  is defined on [0, 2] and can be used to capture the frequency-domain characteristics of different frequency bands in the graph signal.

The node features across different dimensions of actual graph data carry distinct meanings and exhibit varying distributions in the frequency domain. To better capture the frequency-domain information of features across different dimensions, graph filtering is performed separately for each feature of different dimensions.

\begin{equation}
h_{k}^{\left ( c \right ) } = P_{k}^{a,b} \left ( 1-\mathit{\Lambda }  \right )\hat{X}^{\left ( c \right ) }
\end{equation}
where $c$ denotes the feature dimension, $\hat{X}^{\left ( c \right ) } $ represents the $c$-th feature output from the FSTNN module, and $h_{k}^{\left ( c \right ) } \in R^{N\times 1}$ is the output after graph filtering. To comprehensively capture the diverse frequency-domain information, after performing $K$-order graph filtering, the features across $K$ frequency bands are fused via learnable parameters:
\begin{equation}
H^{\left ( c \right ) }  = \sum_{k=0}^{K}\theta _{k}^{\left ( c \right ) }  h_{k}^{\left ( c \right ) }
\end{equation}
where, $\theta _{k}^{\left ( c \right ) }$ is learnable parameter for the $c$-th feature. After applying graph filtering to each feature, the final node representation $H\in R^{N\times d}$ is obtained, which can be used for subsequent graph anomaly detection. To mitigate the class imbalance problem, this module employs a weighted cross-entropy loss function for training:
\begin{equation}
L_{jacobi} = -{\textstyle \sum_{v\in V}^{}} \left [ \varphi y_{v}\mathrm{log} p_{v} +\left ( 1- y_{v}\right )\mathrm{log} \left ( 1- p_{v} \right )    \right ]
\end{equation}
here, $\varphi$ is equal to the ratio of the number of anomalous nodes to the number of normal nodes in the training dataset, and $p_{v}$ denotes the model's predicted probability.

\subsection{Node Label Constraint Module}
To improve node label utilization, this paper introduces a node label constraint module to assist model training. First, FSTNN obtains a more refined node representation $\hat{X}$ by feature transformation. Based on this, it calculates the representation consistency between a central node and its local environment. This is because anomalous nodes typically exhibit greater inconsistency with the features of their surrounding nodes, whereas normal nodes, conversely, tend to show higher consistency. Therefore, the consistency between a node and its neighboring nodes can be used as a criterion to identify anomalous nodes. Therefore, we designed a dimension-wise aggregating network to calculate consistency score for each node:
\begin{equation}
E_{c}^{l}  =\mathrm{GCN}_{c}\left ( E_{c}^{l-1}   \right )
\end{equation}
where, $E_{c}^{l}$ represents the representation of graph node after aggregating the $l$-hop neighborhood information for its $c$-th feature. And the initial input $E_{c}^{0}$ is $\hat{X}$. $\mathrm{GCN}_c()$ is aggregating method from \cite{20KipfW2017GCN} to learn surrounding features. After computing all the features of different dimensions, the local environment node representation $E$ is obtained.

Then, we can compute the consistency score between the central node and its surrounding environment using cosine similarity:
\begin{equation}
s_{i} =cos\left ( \hat{X}_{i},E_{i}  \right )   =\frac{\hat{X}_{i} \cdot E_{i}}{\big\| \hat{X}_{i} \big\| \big\| E_{i} \big\|}
\end{equation}
where $\hat{X}_{i}$ and $E_{i}$ denote the representation and environment representation for node $v_{i}$, respectively. $cos()$ is cosine similarity function. Generally, normal node behavior patterns exhibit high consistency with their local environments, while anomalous nodes often show significant deviations from their neighborhood environments. Therefore, by designing a contrastive constraint loss, we widen the gap in latent node embeddings between normal and anomalous nodes, thereby helping the model more effectively distinguish anomalies. The loss function is formulated as follows:
\begin{equation}
\mathit{L}_{label}= -\log \sigma\!\left(\frac{1}{|v_{nor}|} \sum_{i\in v_{nor}} s_i-\frac{1}{|v_{abn}|} \sum_{i\in v_{abn}} s_i\right)
\end{equation}
where, $v_{nor}$ and $v_{abn}$ represent the index sets for normal samples and anomalous samples, respectively. $\sigma \left ( \cdot  \right )$ denotes the Sigmoid function.

\subsection{Graph Anomaly Detection}
Through the learning of the FSTNN and adaptive Jacobi polynomial graph filtering, its output $H$ contains rich frequency-domain information that can be utilized for graph anomaly detection. During anomaly detection, the output $H$ is first passed through $MLP$ and then normalized using the $softmax$ function to obtain the predicted probability distribution $P\in R^{N\times 2}$:
\begin{equation}
P=softmax\left ( \mathrm{MLP}\left ( H \right )\right )
\end{equation}

During model training, the model is optimized using a weighted combination of the two loss functions:
\begin{equation}
L=L_{jacobi}  +\mu L_{label}
\end{equation}
where $\mu$ serves as the balance weight.

\section{Experiments}
In this section, we perform experiments on five datasets and compare our model's results with those of mainstream approaches to demonstrate the effectiveness of JPGFN.

\subsection{Datasets}
\begin{table}
    \caption{Statistical information of five datasets.}
    \centering
    \setlength{\tabcolsep}{4pt}
    \begin{tabular}{ccccc}
        \hline
        Dataset  & Nodes & Edges & Anomaly(\%) & Features  \\
        \hline
        Amazon     & 11,944   & 4,398,392   & 6.87    & 25 \\
        YelpChi    & 45,954   & 3,846,979   & 14.53   & 32\\
        T-Finance  & 39,357   & 21,222,543  & 4.58    & 10\\
        Elliptic   & 46,564   & 73,248      & 9.76    & 93\\
        Weibo      & 8,405    & 407,963     & 10.33   & 400\\
        \hline
    \end{tabular}
    \label{tab1}
\end{table}
We conduct experiments on five real-world datasets for graph anomaly detection to validate the effectiveness of the proposed method. Detailed information about the datasets is shown in Table~\ref{tab1}. Among them, Amazon \cite{32mcauley2013amazon} aims to identify fraudulent reviews on e-commerce platforms. YelpChi \cite{33rayana2015yelpchi} targets the detection of fraudulent reviews on service platforms. Both T-Finance \cite{13tang2022} and Elliptic \cite{34Elliptic} are utilized to detect fraudulent behaviors in financial transaction networks. The Weibo dataset collected from social platforms \cite{35tang2023gadbench} is used to identify abnormal users in social network.

\subsection{Experimental Metrics}
The performance of the proposed method is evaluated using two widely adopted metrics: AUC-ROC and AUC-PR.

AUC-ROC: This metric quantifies classification performance using the area under the ROC curve. The ROC curve illustrates how the true positive rate (TPR) changes as the false positive rate (FPR) varies. The closer the metric is to 1, the better the model’s classification performance.

AUC-PR: This metric evaluates a model's classification performance by computing the area under the precision–recall (PR) curve, which characterizes the relationship between precision and recall across different decision thresholds. Furthermore, due to the significant imbalance between positive and negative samples in graph anomaly detection tasks, AUC-PR better reflects a model's performance in detecting rare anomaly samples than AUC-ROC; the closer the value is to 1, the better the detection performance.

\subsection{Baseline Methods}
The comparison methods are divided into three categories. The first category consists of common GNNs methods, including GCN \cite{20KipfW2017GCN}, GAT \cite{22Velickovic2018GAT}, GraphSAGE \cite{21hamilton2017GraphSAGE}, and GIN \cite{23xu2018GIN}. The second category consists of approaches based on spatial-domain methods, including GAS \cite{36LiQLYL2019}, PC-GNN \cite{37liu2021}, GDN \cite{10gao2023GDN}, and GFCN \cite{9MesgaranH2024GFCN}. The third category consists of spectral-domain methods, including BWGNN \cite{13tang2022}, AMNet \cite{38chai2022}, SEC-GFD \cite{14XuWWWZW24SECGFD}, AHFAN \cite{7WangDDM2025}, EGNN \cite{liu2026modeling} and DSGAD \cite{15zheng2025DSGAD}.
\begin{enumerate}
    \item[1)] GCN: A traditional GNNs method. This model classifies nodes by aggregating information from first-order neighbors.
    \item[2)] GAT: A traditional GNNs approach. By incorporating an attention mechanism, this model assigns different weights to different nodes when aggregating information from neighboring nodes.
    \item[3)] GraphSAGE: A traditional GNNs method. This model generates new node embeddings for classification by sampling nodes from local neighborhoods and aggregating their features.
    \item[4)] GIN: A traditional GNNs method. This model learns node representations through summation-based aggregation and multi-layer perceptrons, thereby enhancing the model's expressive power.
    \item[5)] GAS: A domain-based anomaly detection method. This model combines homogeneous and heterogeneous graphs to learn the local-global contexts of anomalous objects, respectively.
    \item[6)] PC-GNN: A spatial-based anomaly detection method. By designing subgraph sampling and neighborhood sampling strategies, this model overcomes the class imbalance issue in anomaly detection tasks.
    \item[7)] GDN: A spatial-based anomaly detection method. This model identifies anomalous nodes that deviate significantly from expectations by analyzing the degree of deviation of each node from the normal patterns in its neighborhood.
    \item[8)] GFCN: An spatial-based anomaly detection method. By introducing skip-connected node representations, this model more effectively leverages the structural and attribute information of the graph for anomaly detection.
    \item[9)] BWGNN: A spectral-domain-based anomaly detection method. By incorporating Beta wavelet functions to construct spectral filters, this model more effectively addresses the spectral ``right-shift'' phenomenon caused by anomalous nodes.
    \item[10)] AMNet: A spectral-domain-based anomaly detection method. This model constructs a graph filter using Bernstein polynomials to simultaneously capture both low-frequency and high-frequency signals.
    \item[11)] AHFAN: A spectral-domain-based anomaly detection method. This model proposes a frequency-domain filtering module based on semantic fusion and a node representation module based on attention mechanism to learn effective representations for GAD.
    \item[12)] SEC-GFD: A spectral-domain-based anomaly detection method. To address the issues of heterophilic in graph data and class imbalance, this model proposes hybrid filtering technique and local context constraint for anomaly detection.
    \item[13)] EGNN: A spectral-domain-based anomaly detection method. This method develops graph learning model based on energy-aware mechanism which can learn spectral characteristics via energy-driven feature aggregation.
    \item[14)] DSGAD: A spectral-domain-based anomaly detection method. To address the limitation of traditional wavelet filters which cannot dynamically learn frequency patterns, resulting in incomplete capture of anomaly information. This model proposes a dynamic wavelet filter and a dynamic fusion mechanism for anomaly detection.
\end{enumerate}
\subsection{Implementation Details}
{\begin{table*}[htb]
    \caption{Performance comparison of different methods on five real-life datasets. Boldface indicates the best performance, and underlining indicates the second-best performance.}
    \resizebox{1.0\textwidth}{!}{
    \begin{tabular}{c|cc|cc|cc|cc|cc}
        \hline
        \multicolumn{1}{c|}{\raisebox{-1.2ex}{Model}} & \multicolumn{2}{c|}{Amazon} & \multicolumn{2}{c|}{YelpChi} & \multicolumn{2}{c|}{T-Finance} & \multicolumn{2}{c|}{Elliptic}& \multicolumn{2}{c}{Weibo} \\
        & AUC-ROC & AUC-PR & AUC-ROC & AUC-PR & AUC-ROC & AUC-PR & AUC-ROC & AUC-PR & AUC-ROC & AUC-PR \\
        \hline
        GCN    & $82.39_{\pm 0.41}$ & $35.13_{\pm 1.30}$ & $58.02_{\pm 0.33}$ & $21.33_{\pm 0.37}$ & $92.34_{\pm 2.33}$ & $73.84_{\pm 5.24}$ & $82.17_{\pm 0.59}$ & $21.94_{\pm 2.73}$ & $97.96_{\pm 0.48}$ & $94.4_{\pm 1.04}$\\
        GAT    & $96.66_{\pm 0.95}$ & $86.67_{\pm 1.38}$ & $79.50_{\pm 1.93}$ & $42.41_{\pm 4.55}$ & $92.76_{\pm 1.47}$ & $63.08_{\pm 10.95}$ & $84.44_{\pm 1.73}$ & $25.16_{\pm 4.82}$ & $95.47_{\pm 1.07}$ & $91.06_{\pm 1.53}$\\
        GraphSAGE  & $89.35_{\pm 3.80}$ & $67.50_{\pm 13.19}$ & $84.82_{\pm 1.47}$ & $53.19_{\pm 3.47}$ & $77.63_{\pm 4.46}$ & $30.19_{\pm 11.89}$ & $84.31_{\pm 1.24}$ & $31.43_{\pm 4.54}$ & $93.94_{\pm 1.50}$ & $84.65_{\pm 3.17}$\\
        GIN    & $93.82_{\pm 1.33}$ & $79.22_{\pm 1.51}$ & $77.02_{\pm 0.86}$ & $36.50_{\pm 1.47}$ & $88.07_{\pm 4.13}$ & $63.70_{\pm 4.72}$ & $82.49_{\pm 1.88}$ & $24.73_{\pm 4.52}$ & $95.50_{\pm 0.76}$ & $90.53_{\pm 1.32}$\\
        \hline
        GAS    & $95.90_{\pm 0.95}$ & $85.14_{\pm 1.23}$ & $78.31_{\pm 1.13}$ & $37.55_{\pm 1.97}$ & $92.06_{\pm 3.42}$ & $72.84_{\pm 3.18}$ & $85.24_{\pm 0.96}$ & $49.46_{\pm 5.73}$ & $94.29_{\pm 1.23}$ & $90.32_{\pm 1.43}$ \\
        PC-GNN & $97.02_{\pm 1.15}$ & $87.58_{\pm 1.84}$ & $79.78_{\pm 1.43}$ & $43.79_{\pm 2.47}$ & $92.89_{\pm 1.23}$ & $69.53_{\pm 7.82}$ & $85.19_{\pm 0.76}$ & $40.68_{\pm 2.63}$ & $90.41_{\pm 1.51}$ & $81.93_{\pm 1.83}$\\
        GDN    & $91.54_{\pm 0.81}$ & $84.88_{\pm 0.89}$ & $79.48_{\pm 0.73}$ & $44.26_{\pm 2.23}$ & $93.41_{\pm 0.78}$ & $71.98_{\pm 2.79}$ & $85.90_{\pm 0.38}$ & $67.62_{\pm 4.17}$ & $91.65_{\pm 0.69}$ & $84.88_{\pm 1.14}$\\
        GFCN   & $94.75_{\pm 0.31}$ & $79.18_{\pm 1.15}$ & $76.43_{\pm 0.09}$ & $40.62_{\pm 0.29}$ & $89.42_{\pm 0.49}$ & $59.89_{\pm 1.39}$ & $82.33_{\pm 1.65}$ & $48.81_{\pm 2.64}$ & $96.85_{\pm 0.03}$ & $91.22_{\pm 0.18}$\\
        \hline
        BWGNN  & $96.44_{\pm 2.10}$ & $87.79_{\pm 2.54}$ & $84.62_{\pm 0.91}$ & $54.84_{\pm 2.10}$ & $95.22_{\pm 0.41}$ & $81.87_{\pm 3.27}$ & $86.92_{\pm 1.33}$ & $59.22_{\pm 6.32}$ & $96.88_{\pm 0.27}$ & $92.60_{\pm 0.76}$\\
        AMNet  & $97.17_{\pm 1.67}$ & $87.61_{\pm 2.19}$ & $82.60_{\pm 0.51}$ & $48.99_{\pm 1.25}$ & $93.58_{\pm 0.84}$ & $74.94_{\pm 2.91}$ & $87.28_{\pm 3.45}$ & $67.86_{\pm 2.43}$ & $95.12_{\pm 1.53}$ & $90.22_{\pm 2.14}$\\
        SEC-GFD   & $97.13_{\pm 0.67}$ & $86.13_{\pm 1.20}$ & $86.24_{\pm 0.47}$ & $59.73_{\pm 1.25}$ & $95.43_{\pm 0.37}$ & $84.09_{\pm 1.89}$ & $88.50_{\pm 0.47}$ & $69.69_{\pm 3.47}$ & $96.01_{\pm 0.55}$ & $90.65_{\pm 0.82}$\\
        AHFAN  & $96.34_{\pm 0.81}$ & $82.82_{\pm 1.05}$ & $85.93_{\pm 0.59}$ & $61.17_{\pm 1.32}$ & $92.78_{\pm 0.11}$ & $77.88_{\pm 0.46}$ & $87.47_{\pm 0.33}$ & $70.89_{\pm 1.74}$ & $97.64_{\pm 0.33}$ & $92.82_{\pm 0.66}$\\
        EGNN  & $95.55_{\pm 1.07}$ & \underline{$88.68_{\pm 1.25}$} & \underline{$89.33_{\pm 0.48}$} & \underline{$68.53_{\pm 0.85}$} & $95.44_{\pm 0.44}$ & $84.13_{\pm 0.93}$ & $85.02_{\pm 4.69}$ & $42.16_{\pm 10.16}$ & $98.17_{\pm 0.73}$ & $93.82_{\pm 1.37}$\\
        DSGAD   & \underline{$97.65_{\pm 0.38}$} & $88.43_{\pm 1.40}$ & $85.99_{\pm 0.61}$ & $60.34_{\pm 0.75}$ & \underline{$96.46_{\pm 0.14}$} & \underline{$86.59_{\pm 0.42}$} & \underline{$88.77_{\pm 1.10}$} & \underline{$71.59_{\pm 2.08}$} & \underline{$98.23_{\pm 0.47}$} & \underline{$93.83_{\pm 0.44}$}\\
        \hline
        JPGFN   & $\mathbf{98.40}_{\mathbf{\pm 0.27}}$ & $\mathbf{90.45}_{\mathbf{\pm 0.69}}$ & $\mathbf{92.09}_{\mathbf{\pm 0.30}}$ & $\mathbf{75.02}_{\mathbf{\pm 0.71}}$ & $\mathbf{96.78}_{\mathbf{\pm 0.12}}$ & $\mathbf{86.84}_{\mathbf{\pm 0.43}}$ & $\mathbf{92.45}_{\mathbf{\pm 0.83}}$ & $\mathbf{77.18}_{\mathbf{\pm 1.47}}$ & $\mathbf{98.98}_{\mathbf{\pm 0.26}}$ & $\mathbf{95.54}_{\mathbf{\pm 0.57}}$ \\
        \hline
    \end{tabular}
    }
    \label{tab2}
\end{table*}
All experiments were conducted on a Tesla A40-48G GPU and an Intel Xeon Gold 6326 CPU. Furthermore, to ensure fairness, all experiments were conducted under identical conditions.

For the baseline models, those based on traditional GNNs methods (GCN, GAT, GraphSAGE and GIN) utilized functions from python library. For the other GAD approaches (PC-GNN, GAS, GDN, BWGNN, GFCN, AMNet, AHFAN, SEC-GFD, EGNN and DSGAD), experiments are conducted utilizing code provided by the original authors. In experiments, the results are obtained by running each model 10 times independently on each dataset and taking the mean as the comparison result, while also recording the standard deviation.

In the experiments, the Optuna framework \cite{39akiba2019optuna} is employed to conduct parameter grid search and obtain the best results. To ensure fair comparisons, we adopt the most commonly used dataset splitting ratios from \cite{35tang2023gadbench}. The Amazon and T-Finance datasets use training/test/validation ratio of 4/4/2, the YelpChi dataset uses 7/2/1, the Elliptic dataset uses 4.5/3.5/2, and the Weibo dataset uses 6/3/1. All experiments are carried out under the same experimental settings. For each dataset, we conduct ten independent experiments and use their average values as the comparative results. The training configurations for the datasets are as follows: Amazon, YelpChi, and T-Finance are trained with learning rate of 0.01 and hidden layer dimension of 64; Elliptic uses a learning rate of 0.05 with a hidden layer dimension of 16; and Weibo is trained with a learning rate of 0.01 and a hidden layer dimension of 32. The maximum order of the graph filter is set to 4 to ensure the filter can learn features from different frequency bands.

\subsection{Performance Evaluation}
Table~\ref{tab2} summarizes the experimental results. For each column, the best result is highlighted in bold, and the second-best result is indicated by underline. As can be seen from the data in the table, compared to the baseline method, the JPGFN consistently achieves the highest anomaly detection accuracy across all five datasets. This demonstrates that the model possesses good generalization capabilities and stronger graph anomaly detection performance. Specifically, JPGFN achieves improvements of 0.75\% and 1.77\% in AUC-ROC and AUC-PR metrics on Amazon, 2.76\% and 6.49\% on YelpChi, on T-Finance by 0.32\% and 0.25\%, on Elliptic by 3.68\% and 5.59\%, and on Weibo by 0.75\% and 1.71\%.

The results in the table also show that the four general GNNs models exhibit poor anomaly detection performance, whereas the rest of GNNs-based methods demonstrate superior performance. The fundamental reason lies in the fact that general GNNs rely on the assumption of homogeneity, while neglecting the issue of heterophily which is critical in anomaly detection scenarios. This causes the models to perform low-pass filtering when aggregating neighborhood information, thereby smoothing out the high-frequency signals of anomalous nodes and limiting the models' anomaly detection performance. Furthermore, spectral-domain-based methods perform slightly better than spatial-domain-based methods, as the latter cannot analyze the signals of anomalous nodes from a frequency-domain perspective. Spectral-domain-based methods employ various strategies to design filters that learn the frequency-domain features of the dataset, enabling them to capture the high-frequency signals generated by anomalous nodes and thus achieve better results.

A comparison with five spectral-domain methods shows that the JPGFN model achieves better performance. This is because all five spectral-domain methods construct graph filters using fixed basis functions. For example, the BWGNN, SEC-GFD, and DSGAD models use Beta wavelet functions, AMNet uses Bernstein polynomials, and AHFAN uses Chebyshev polynomials to construct graph filters. These fixed basis functions lack flexibility, whereas the frequency-domain distributions of real-world graph datasets are diverse, preventing them from effectively learning frequency-domain information. Furthermore, these methods do not fully account for the impact of each attribute of graph node on GAD. The JPGFN introduces FSTNN and adaptive Jacobi polynomial graph filtering to effectively address these two issues. It also makes better use of node labels to assist in model training, thereby significantly improving the performance of graph anomaly detection.

\subsection{Ablation Study}
\label{ablation}
\begin{table*}[htb]
    \caption{The ablation study of JPGFN. Here, w/o denotes without this module and w/ denotes with this module}
\resizebox{1.0\linewidth}{!}{
    \begin{tabular}{ccccccccccc}
        \hline
        \multicolumn{1}{c}{\raisebox{-1.2ex}{Model}} & \multicolumn{2}{c}{Amazon} & \multicolumn{2}{c}{YelpChi} & \multicolumn{2}{c}{T-Finance} & \multicolumn{2}{c}{Elliptic} & \multicolumn{2}{c}{Weibo} \\
        & AUC-ROC & AUC-PR & AUC-ROC & AUC-PR & AUC-ROC & AUC-PR & AUC-ROC & AUC-PR & AUC-ROC & AUC-PR\\
        \hline
        w/o FSTNN    & $95.54_{\pm 1.26}$ & $84.70_{\pm 2.16}$ & $82.82_{\pm 0.99}$ & $52.12_{\pm 3.04}$ & $92.78_{\pm 1.22}$ & $73.16_{\pm 4.89}$ & $87.41_{\pm 0.47}$ & $61.39_{\pm 6.23}$ & $96.18_{\pm 0.55}$ & $91.87_{\pm 0.63}$\\
        w/ MLP       & $98.20_{\pm 0.17}$ & $89.17_{\pm 0.56}$ & $86.06_{\pm 0.29}$ & $60.01_{\pm 0.64}$ & $95.19_{\pm 0.25}$ & $82.21_{\pm 1.12}$ & $88.28_{\pm 0.41}$ & $67.21_{\pm 3.59}$ & $98.27_{\pm 0.18}$ & $94.78_{\pm 0.59}$\\
        \hline
        w/o Jacobi   & $97.60_{\pm 0.41}$ & $88.43_{\pm 1.08}$ & $89.58_{\pm 0.33}$ & $67.88_{\pm 1.17}$ & $93.51_{\pm 0.19}$ & $74.33_{\pm 0.63}$ & $92.05_{\pm 1.39}$ & $75.93_{\pm 2.29}$ & $95.90_{\pm 0.55}$ & $88.51_{\pm 0.72}$\\
        w/ Chebyshev & $98.07_{\pm 0.31}$ & $89.35_{\pm 0.90}$ & $89.60_{\pm 0.24}$ & $68.70_{\pm 0.54}$ & $96.33_{\pm 0.12}$ & $84.76_{\pm 0.69}$ & $87.16_{\pm 1.69}$ & $62.83_{\pm 4.76}$ & $\mathbf{98.99}_{\mathbf{\pm 0.33}}$ & $95.47_{\pm 0.55}$\\
        w/ Legendre  & $98.28_{\pm 0.22}$ & $89.55_{\pm 0.59}$ & $91.71_{\pm 0.33}$ & $73.73_{\pm 0.74}$ & $96.65_{\pm 0.20}$ & $86.25_{\pm 0.36}$ & $88.96_{\pm 2.09}$ & $67.30_{\pm 4.06}$ & $98.73_{\pm 0.42}$ & $95.05_{\pm 0.40}$\\
        w/ Bernstein & $98.28_{\pm 0.16}$ & $90.07_{\pm 0.59}$ & $91.50_{\pm 0.36}$ & $73.60_{\pm 0.74}$ & $96.34_{\pm 0.10}$ & $85.19_{\pm 0.38}$ & $90.96_{\pm 1.16}$ & $74.49_{\pm 1.42}$ & $97.02_{\pm 0.38}$ & $91.82_{\pm 0.67}$\\
        \hline
        w/o Label & $98.33_{\pm 0.28}$ & $89.88_{\pm 0.59}$ & $91.58_{\pm 0.23}$ & $74.24_{\pm 0.50}$ & $96.75_{\pm 0.22}$ & $86.70_{\pm 0.34}$ & $92.15_{\pm 1.27}$ & $76.59_{\pm 1.58}$ & $\mathbf{98.99}_{\mathbf{\pm 0.30}}$ & $95.44_{\pm 0.43}$\\
        \hline
        JPGFN   & $\mathbf{98.40}_{\mathbf{\pm 0.27}}$ & $\mathbf{90.45}_{\mathbf{\pm 0.69}}$ & $\mathbf{92.09}_{\mathbf{\pm 0.30}}$ & $\mathbf{75.02}_{\mathbf{\pm 0.71}}$ & $\mathbf{96.78}_{\mathbf{\pm 0.12}}$ & $\mathbf{86.84}_{\mathbf{\pm 0.43}}$ & $\mathbf{92.45}_{\mathbf{\pm 0.83}}$ & $\mathbf{77.18}_{\mathbf{\pm 1.47}}$ & $98.98_{\pm 0.26}$ & $\mathbf{95.54}_{\mathbf{\pm 0.57}}$ \\
        \hline
    \end{tabular}
    \label{tab3}
    }
\end{table*}
Here, we validate the importance of the three modules by removing each module individually as follows, producing the corresponding variants as shown in parentheses: FSTNN (w/o FSTNN), adaptive Jacobi polynomial graph filtering module(w/o Jacobi), and the node label constraint module (w/o Label) through ablation experiments. For FSTNN, we replaced it with a  MLP producing variant w/ MLP. For the adaptive Jacobi polynomial graph filtering module, we designed three variant methods, substituting the Jacobi polynomial with three other commonly used polynomials: Chebyshev polynomial(w/ Chebyshev), Legendre polynomial(w/ Legendre), Bernstein polynomial(w/ Bernstein). In the experimental setup, all components and parameter settings remained unchanged except for the replaced module. Each variant model is run independently 10 times on each of the five datasets, and the mean and standard deviation of AUC-ROC and AUC-PR are recorded.

Table~\ref{tab3} presents the experimental results across five datasets. The results show that JPGFN achieved the best performance, confirming the effectiveness of the proposed module. First, compared to w/o FSTNN, the AUC-ROC of the JPGFN model improved by 2.86\%, 9.27\%, 4\%, 5.04\%, and 2.8\% across the five datasets, and the AUC-PR metrics improved by 5.75\%, 22.9\%, 13.68\%, 15.79\%, and 3.67\%, respectively, demonstrating the effectiveness of FSTNN. Compared to the variant using an MLP (i.e., w/ MLP), the AUC-ROC of JPGFN improved by 0.2\%, 6.03\%, 1.59\%, 4.17\%, and 0.71\% across the five datasets, and AUC-PR metrics improved by 1.28\%, 15.01\%, 4.63\%, 9.97\%, and 0.76\%, respectively. This indicates that it is essential to account for the varying importance of different node features in graph anomaly detection tasks.

Secondly, compared to the w/o Jacobi model, the JPGFN model achieved AUC-ROC improvements of 0.8\%, 2.51\%, 3.27\%, 0.4\%, and 3.08\%, and the AUC-PR metrics improved by 2.02\%, 7.14\%, 12.51\%, 1.25\%, and 7.03\%, respectively, demonstrating the effectiveness of the adaptive Jacobi polynomial graph filtering module. Furthermore, when comparing the three polynomial variants, the JPGFN model's AUC-ROC and AUC-PR metrics are consistently improved, validating the superiority of the Jacobi polynomials. This indicates that Jacobi polynomials offer greater flexibility and can adapt to the frequency characteristics of different graph datasets. In contrast, Chebyshev polynomials, Legendre polynomials, and Bernstein polynomials, due to their fixed basis function forms, perform well on certain datasets but exhibit significantly insufficient overall adaptability.

Finally, compared to the w/o Label approach, the JPGFN model achieved AUC-ROC improvements of 0.07\%, 0.51\%, 0.03\%, 0.3\%, and -0.01\%, and the AUC-PR metrics improved by 0.57\%, 0.78\%, 0.14\%, 0.59\%, and 0.1\%, respectively, demonstrating the effectiveness of the node label constraint module. The experimental results demonstrate that incorporating the label constraint module leads to an overall improvement in the model's performance.

\subsection{Parameter Analysis}
\begin{figure*}[!t]
\centering
\includegraphics[width=0.90\textwidth]{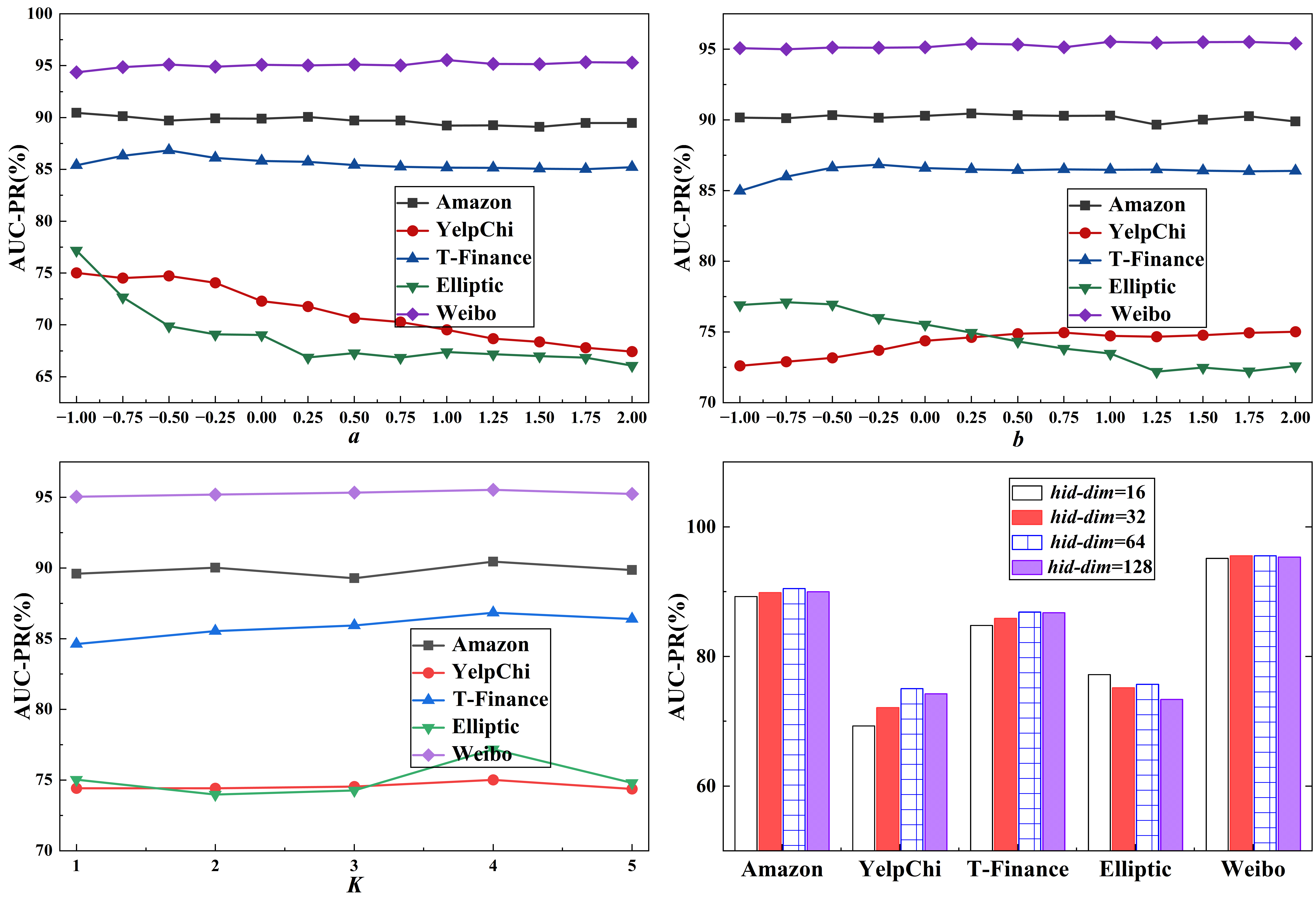}
\caption{Experimental results of parameter analysis. The figure shows the trends of the AUC-PR metric changing with the parameters $a$, $b$, the polynomial order $K$, and the hidden layer dimension $hid\text{-}dim$.}
\label{fig4}
\end{figure*}
This section provides parameter analysis of the Jacobi polynomial parameters $a$ and $b$, the order $K$, and the hidden layer dimension $hid\text{-}dim$. In the adaptive Jacobi polynomial graph filter module, the parameters $a$ and $b$ can determine the learning functions of graph filters. In the model, they are parameterized as learnable parameters to adaptive learn complex spectral distribution of different graph data. To facilitate the parameter analysis, we manually set these parameters to different values, thereby enabling a systematic evaluation of their effects on the model performance. The order $K$ of the graph filter determines the frequency range that the graph filter can learn. The hidden layer dimension $hid\text{-}dim$ determines the representational capacity of the model in the high-dimensional latent space. Figure~\ref{fig4} illustrates the trends of model performance across five datasets as parameters vary.

As can be seen from the trends in the top two scatter plots in Figure~\ref{fig4}, different values of $a$ and $b$ have significant impact on performance for the Elliptic and Yelp datasets, while performance also varies with changes in $a$ and $b$ for the other three datasets. The optimal parameter values differ across datasets. For the Elliptic dataset, performance is optimal when $a$ and $b$ are -1.0 and 2.0, respectively; for the YelpChi dataset, they are -0.75 and -0.5; for the Amazon dataset, they are -1.0 and 0.25; for the T-Finance dataset, they are -0.5 and -0.25; and for the Weibo dataset, they are 1.0 and 1.0. This confirms that, in the proposed model, adaptive learning of different $a$ and $b$ values for different datasets can results in better adapt to the frequency-domain distribution of different datasets.

The scatter plot at the bottom left of Figure~\ref{fig4} illustrates the impact of the polynomial order $K$ on model performance. As shown in the figure, the performance of the model exhibits different trends with respect to the value of $K$ across different datasets. This is because different graph datasets exhibit distinct frequency-domain distributions, leading to different underlying frequency-domain patterns. Therefore, different values of $K$ should be adopted to more effectively capture and learn the frequency-domain patterns inherent in different graph datasets.

The bar chart at the bottom right of Figure~\ref{fig4} illustrates the impact of the hidden layer dimension on performance. It can be observed that the best results are achieved with a hidden layer dimension of 64 on the Amazon, YelpChi, and T-Finance datasets; with a hidden layer dimension of 16 on the Elliptic dataset; and with a hidden layer dimension of 32 on the Weibo dataset. This indicates that different datasets require different hidden layer dimensions, which are dictated by the specific characteristics of each graph dataset. Too few dimensions fail to adequately capture the node features and frequency-domain information of the graph data, while too many dimensions can lead to overfitting. Therefore, an appropriate hidden layer dimension must be set for each dataset.

\subsection{Visualization}
\begin{figure}
    \centering
    % 第一行
    \begin{subfigure}{0.22\textwidth}
        \centering
        \includegraphics[width=\linewidth]{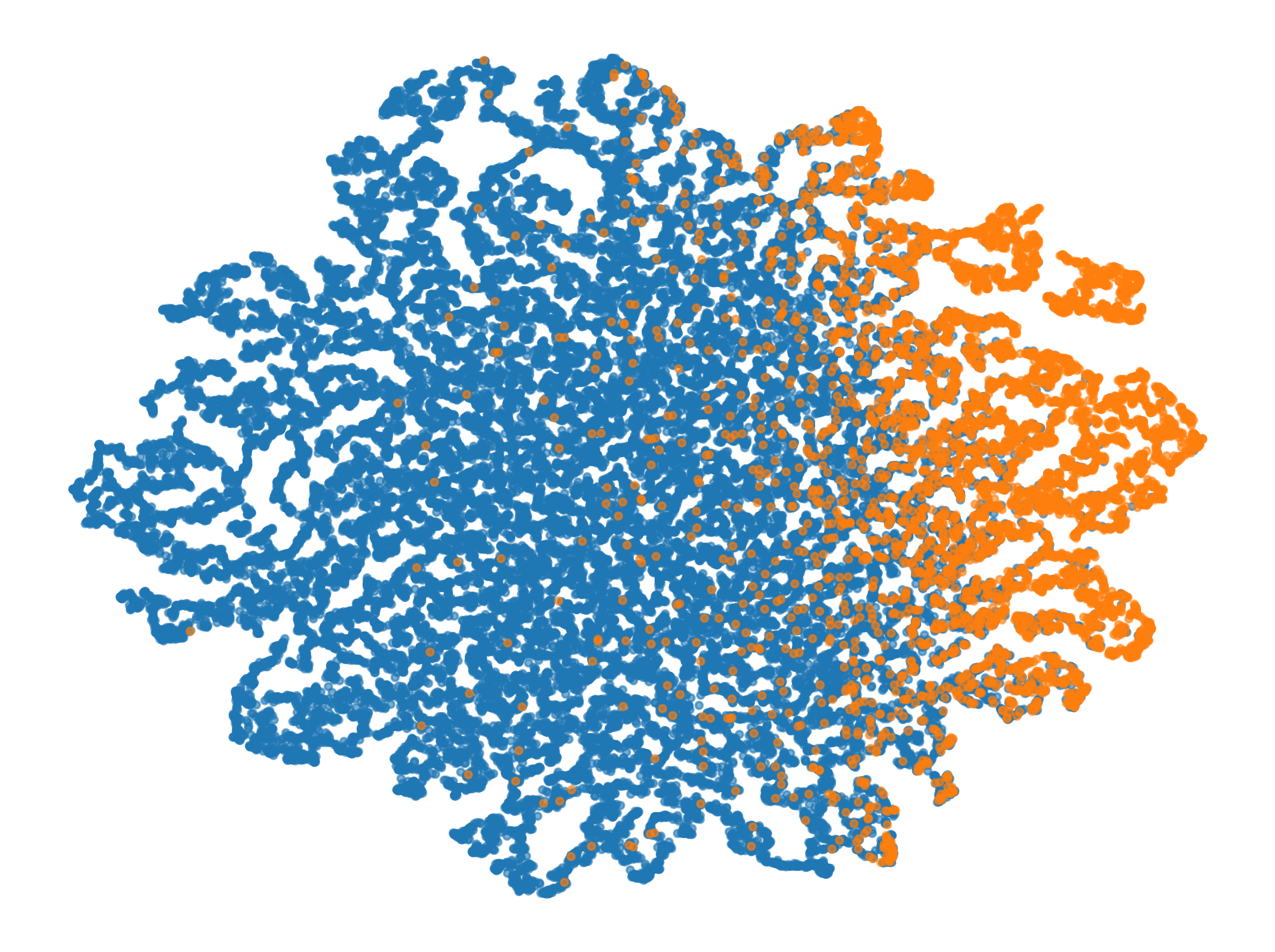}
        \caption{JPGFN}
    \end{subfigure}
    \hfill
    \begin{subfigure}{0.22\textwidth}
        \centering
        \includegraphics[width=\linewidth]{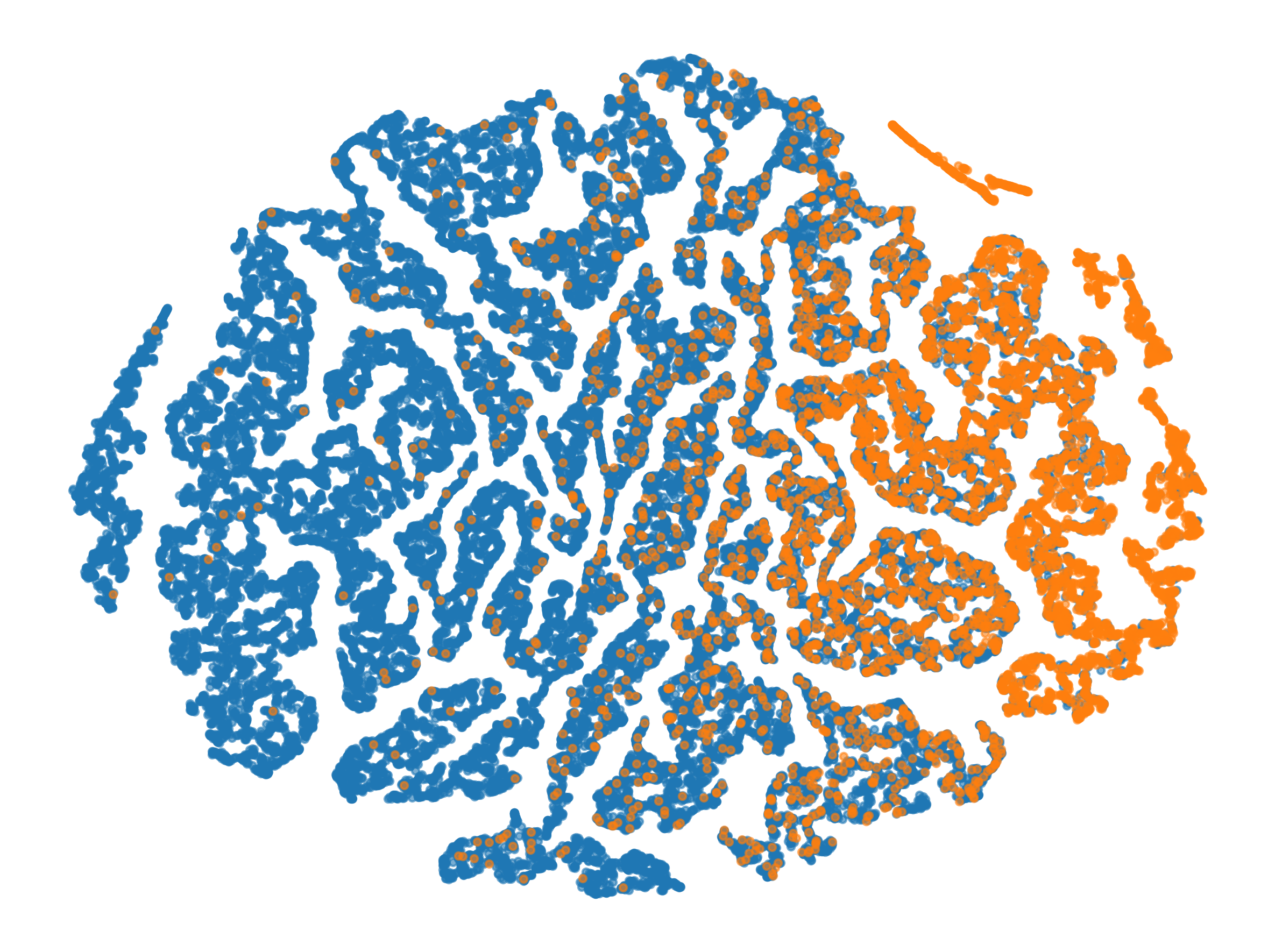}
        \caption{w/ MLP}
    \end{subfigure}
    \vspace{0.3cm}
    % 第二行
    \begin{subfigure}{0.22\textwidth}
        \centering
        \includegraphics[width=\linewidth]{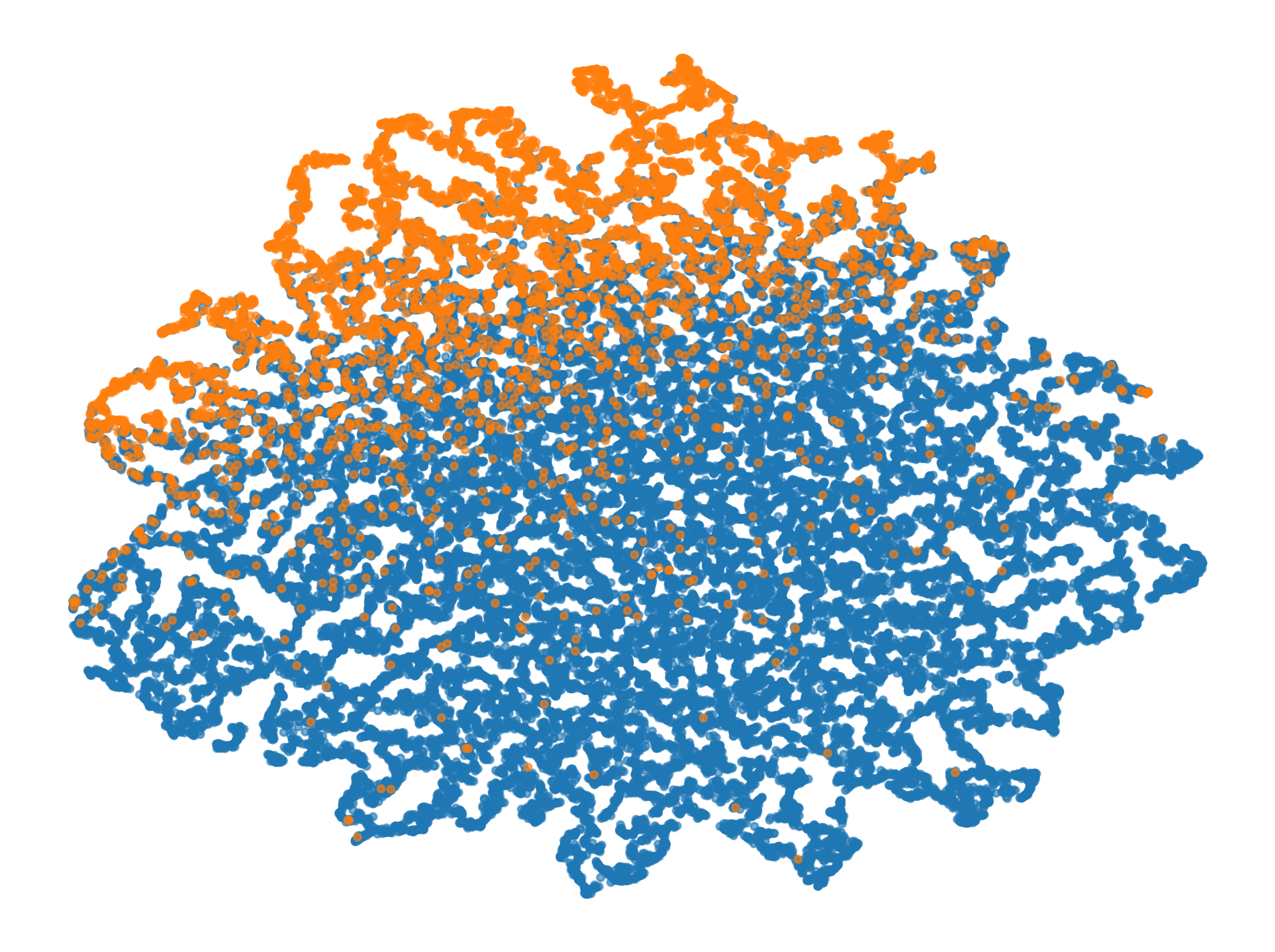}
        \caption{w/o Jacobi}
    \end{subfigure}
    \hfill
    \begin{subfigure}{0.22\textwidth}
        \centering
        \includegraphics[width=\linewidth]{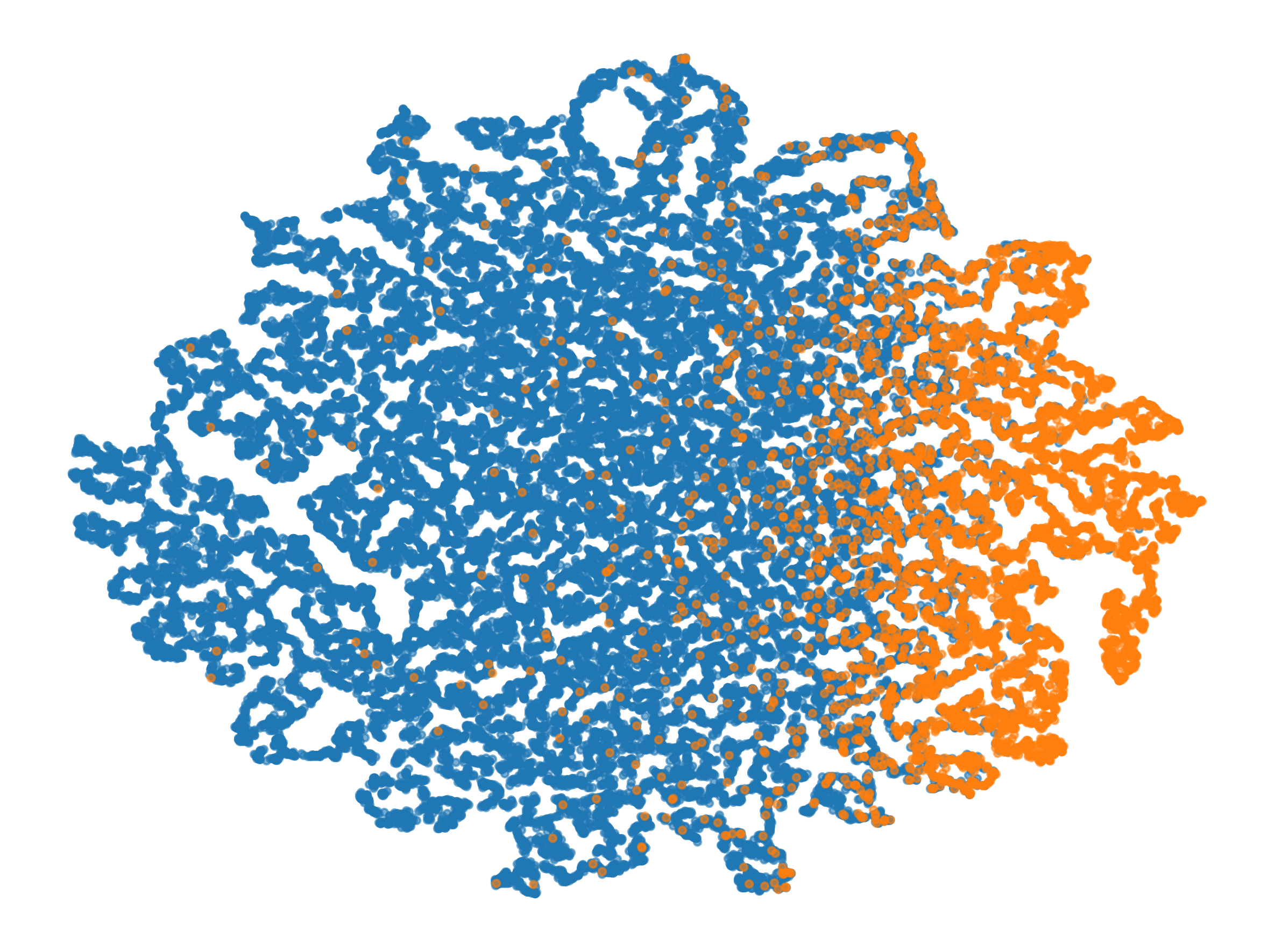}
        \caption{w/o Label}
    \end{subfigure}
    \caption{Visualization results on YelpChi.}
    \label{fig5}
\end{figure}
To more intuitively analyze the effectiveness of the three modules in the JPGFN model, we performed visual analysis of the node representations generated by the JPGFN model and the three variant methods described in Section \ref{ablation}. For the visualization study, the YelpChi dataset was used to conduct the experiments. Specifically, we used the t-SNE \cite{40maaten2008visualizing} dimension reduction method to map the model's node representations into a two-dimensional space, labeling node categories with different colors: blue for normal nodes and yellow for abnormal nodes. The visualization results are shown in Figure~\ref{fig5}.

As observed, the visualization quality w/ MLP is relatively poor, primarily due to insufficient consideration of the importance of each attribute within node features. The visualization quality of w/o Jacobi is also inferior to JPGFN, as the Jacobi polynomial graph filter module can learn rich frequency domain information. The visualization of w/o Label shows inferior visual effect compared to JPGFN. This is because removing the node label constrain module underutilizes the label information and reduces model performance. JPGFN demonstrates the best visualization effect, showing more distinct separation between normal and anomalous nodes in the visualization space.

\subsection{Analysis of Time Complexity}
To further evaluate the scalability and computational efficiency of the proposed JPGFN model, this section presents an analysis of its theoretical complexity and empirical execution time. First, the complexity of FSTNN is $Nnd^{2}$ , where $n$ is the feature dimension of the nodes and $d$ is the dimension of the hidden layer. The complexity of the Jacobi polynomial graph filtering module is $KEd$, where $K$ is the polynomial order and $E$ is the number of edges. The complexity of the node label constraint module is $O\left ( E+Nd^{2}  \right )$ . Therefore, the theoretical complexity of the JPGFN model is $Nnd^{2}+KEd$. Since $n$ is small, the complexity of FSTNN is comparable to that of MLP. The adaptive Jacobi polynomial graph filtering module belongs to the category of $K$th-order polynomial graph filters, and its time complexity is comparable to that of a general polynomial graph filter. The node label constraint module processes features through a combination of graph convolution and linear operations, and its impact on the overall complexity is the same as graph convolution.

Since actual runtime depends not only on theoretical complexity but also on the coefficients and parameters used in the model. We conducted experiments on the YelpChi dataset to evaluate actual runtime. Table~\ref{tab4} shows the actual training cost and inference cost for the JPGFN model and they compare with four other state-of-the-art graph anomaly detection methods. Training time refers to the average time consumed per training round on the training set, whereas inference time refers to the time required to perform prediction on the test set. Table~\ref{tab4} shows that JPGFN requires less computational time than AHFAN. But compared to SEC-GFD and DSGAD, JPGFN takes longer time for training and inference. This is because AHFAN employs attention mechanism to learn node representation, which results in longer processing times of graph data. And for JPGFN model, the introduction of feature separation operation can increase computational overhead.

\begin{table}[h]
    \caption{Comparison of Time Complexity for Different Methods.}
    \centering
    % \small
    \setlength{\tabcolsep}{4pt}
    \begin{tabular}{lcccc}
        \hline
        \multicolumn{1}{l}{\multirow{3}{*}{Model}} & \multicolumn{2}{c}{Calculation time} & \multicolumn{1}{c}{\multirow{3}{*}{Time complexity}} & \\
        \cline{2-3}
        & \shortstack{Training \\[-2pt] (ms/epoch)} & \shortstack{Inference \\[-2pt] (ms)} & \\
        \hline
        GAT       & 96     & 42    & $O\left ( MNd^{2} +MEd \right )$    \\
        AHFAN     & 197    & 94    & $O\left ( Nd^{2} + Ed \right )$     \\
        SEC-GFD   & 42     & 78   & $O\left ( Nd^{2} + K^{2}Ed \right )$ \\
        DSGAD     & 61     & 32    & $O\left ( LNd^{2} + LKEd \right )$  \\
        JPGFN     & 116    & 95    & $O\left ( Nnd^{2} +KEd \right )$    \\
        \hline
    \end{tabular}
    \label{tab4}
\end{table}

\section{Conclusion and Future Work}
This work proposes a novel GAD framework, JPGFN, which effectively identifies anomalous nodes in graphs. First, by introducing the Feature Separation Transformation Network (FSTNN), it better learns fine-grained information of node features. Second, we construct adaptive graph filters using Jacobi polynomials and its parameterized form, achieving more flexible learning of frequency-domain information. Finally, we construct a node label constraint module which improves the model's performance by incorporating node labels for training. Experimental outcomes on five real-life datasets validate the superiority of the proposed method over existing mainstream baseline methods. Future work will explore designing more effective graph filters for learning frequency-domain information in graph signals and extending the method to more complex scenarios such as dynamic and heterogeneous graphs.

% Numbered list
% Use the style of numbering in square brackets.
% If nothing is used, default style will be taken.
%\begin{enumerate}[a)]
%\item
%\item
%\item
%\end{enumerate}

% Unnumbered list
%\begin{itemize}
%\item
%\item
%\item
%\end{itemize}

% Description list
%\begin{description}
%\item[]
%\item[]
%\item[]
%\end{description}

% \clearpage %%Remove this from your manuscript

% Uncomment and use as the case may be
%\begin{theorem}
%\end{theorem}

% Uncomment and use as the case may be
%\begin{lemma}
%\end{lemma}

%% The Appendices part is started with the command \appendix;
%% appendix sections are then done as normal sections
%% \appendix

% \section{}\label{}

% To print the credit authorship contribution details
%%\printcredits
%% Loading bibliography style file
%\bibliographystyle{model1-num-names}
\bibliographystyle{elsarticle-num}

% Loading bibliography database
\bibliography{refs}

@article{LIU2026112935,
title = {Geometric and topological structure-induced large-scale graph learning for social and information networks},
journal = {Pattern Recognition},
volume = {173},
pages = {112935},
year = {2026},
author = {Gang Hao Liu and Ting Xiao and Zhe Wang and others}
}

@article{2motie2024,
  title={{Financial fraud detection using graph neural networks: A systematic review}},
  author={Motie, Soroor and Raahemi, Bijan},
  journal={Expert Systems with Applications},
  volume={240},
  pages={122156},
  year={2024}
}

@article{3cui2025,
  title={{Enhancing link prediction in biomedical knowledge graphs with BioPathNet}},
  author={Emy Yue Hu and Svitlana Oleshko and Samuele Firmani and others},
  journal={Nature Biomedical Engineering},
  volume={ },
  number={ },
  pages={ },
  doi={https://doi.org/10.1038/s41551-025-01598-z},
  year={2026}
}

@article{4Huang2025,
  author       = {Hao Qi Huang and
                  Ping Wang and
                  Jian Hua Pei and others},
  title        = {{Deep Learning Advancements in Anomaly Detection: A Comprehensive
                  Survey}},
  journal      = {IEEE Internet of Things Journal},
  volume       = {12},
  number       = {21},
  pages        = {44318--44342},
  year         = {2025}
}

@article{5HojjatiHA2024,
  author       = {Hadi Hojjati and
                  Thi Kieu Khanh Ho and
                  Narges Armanfard},
  title        = {Self-supervised anomaly detection in computer vision and beyond: {A}
                  survey and outlook},
  journal      = {Neural Networks},
  volume       = {172},
  pages        = {106106},
  year         = {2024}
}

@article{HEVAPATHIGE2026108869,
title = {{Permutation-Invariant graph partitioning: How graph neural networks capture structural interactions?}},
journal = {Neural Networks},
volume = {200},
pages = {108869},
year = {2026},
author = {Asela Hevapathige and Qing Wang}
}

@article{7WangDDM2025,
  author       = {Xiang Wang and
                  Hao Dou and
                  Dibo Dong and
                  others},
  title        = {Graph anomaly detection based on hybrid node representation learning},
  journal      = {Neural Networks},
  volume       = {185},
  pages        = {107169},
  year         = {2025}
}

@article{9MesgaranH2024GFCN,
  author       = {Mahsa Mesgaran and
                  A. Ben Hamza},
  title        = {Graph fairing convolutional networks for anomaly detection},
  journal      = {Pattern Recognition},
  volume       = {145},
  pages        = {109960},
  year         = {2024}
}

@inproceedings{10gao2023GDN,
  title={Alleviating structural distribution shift in graph anomaly detection},
  author={Gao, Yuan and Wang, Xiang and He, Xiangnan and Liu, Zhenguang and Feng, Huamin and Zhang, Yongdong},
  booktitle={Proceedings of the ACM International Conference on Web Search and Data Mining},
  pages={357--365},
  year={2023}
}

@article{11zhang2025,
  author={Liu, Zhi Zhe and Zheng, Shuai and Yan, Ye Yu and others},
  journal={IEEE Transactions on Network Science and Engineering},
  title={{Adaptive Graph Filtering Neural Network for Graph Anomaly Detection}},
  year={2026},
  volume={13},
  number={},
  pages={3274--3284}
  }

@article{12ding2025novel,
  title={{SPS-GAD: Spectral-spatial graph structure learning for anomaly detection in heterophilic graphs}},
  author={Chen Zhu, Ya Ying Zhang},
  journal={Expert Systems With Applications},
  volume={298},
  number={ },
  pages={129639}, 
  year={2026}
}

@inproceedings{13tang2022,
  title={Rethinking graph neural networks for anomaly detection},
  author={Tang, Jian Heng and Li, Jia Jin and Gao, Zi Qi and others},
  booktitle={Proceedings of the International Conference on Machine Learning},
  pages={21076--21089},
  year={2022}
}

@inproceedings{14XuWWWZW24SECGFD,
  title={Revisiting graph-based fraud detection in sight of heterophily and spectrum},
  author={Xu, Fan and Wang, Nan and Wu, Hao and others},
  booktitle={Proceedings of the AAAI Conference on Artificial Intelligence},
  pages={9214--9222},
  year={2024}
}

@inproceedings{15zheng2025DSGAD,
  title={Dynamic Spectral Graph Anomaly Detection},
  author={Zheng, Jian Bo and Yang, Chao and Zhang, Tai Rui and others},
  booktitle={Proceedings of the AAAI Conference on Artificial Intelligence},
  pages={13410--13418},
  year={2025}
}

@article{16eliasof2024global,
title = {Adaptive message passing mechanism for graph neural networks},
journal = {Pattern Recognition},
volume = {179},
pages = {113875},
year = {2026},
author = {Yang Tao Wang and Yu Jie Shi and Qi Zhang and others}}

@article{17ratna2025inclusive,
  title={An inclusive analysis for performance and efficiency of graph neural network models for node classification},
  author={Ratna, S and Singh, Sukhdeep and Sharma, Anuj},
  journal={Computer Science Review},
  volume={56},
  pages={100722},
  year={2025}
}

@inproceedings{18wang2024graph,
  title={Graph classification via reference distribution learning: theory and practice},
  author={Wang, Zixiao and Fan, Jicong},
  booktitle = {Proceedings of the Advances in Neural Information Processing Systems},
  pages={137698--137740},
  year={2024}
}

@inproceedings{19qian2025exploring,
  title={Exploring the over-smoothing problem of graph neural networks for graph classification: an entropy-based viewpoint},
  author={Qian, Fei Fei and Bai, Lu and Cui, Li Xin and others},
  booktitle={Proceedings of the AAAI Conference on Artificial Intelligence},
  pages={19995--20003},
  year={2025}
}

@inproceedings{20KipfW2017GCN,
  title={{Semi-Supervised Classification with Graph Convolutional Networks}},
  author={Kipf, Thomas N and Welling, Max},
  booktitle={Proceedings of the International Conference on Learning Representations},
  pages = {1--14},
  year={2017}
  }

@inproceedings{21hamilton2017GraphSAGE,
  title={Inductive representation learning on large graphs},
  author={Hamilton, Will and Ying, Zhi Tao and Leskovec, Jure},
  booktitle = {Proceedings of the Advances in Neural Information Processing Systems},
  pages = {1024--1034},
  year={2017}
}

@inproceedings{22Velickovic2018GAT,
  title={Graph attention networks},
  author={Velickovic, Petar and Cucurull, Guillem and Casanova, Arantxa and others},
  booktitle={Proceedings of the International Conference on Learning Representations},
  pages={1--12},
  year={2018}
}

@inproceedings{23xu2018GIN,
title={{How Powerful are Graph Neural Networks?}},
author={Keyulu Xu and Wei Hua Hu and Jure Leskovec and others},
booktitle={Proceedings of the International Conference on Learning Representations},
year={2019}
}

@article{24he2024polarized,
  title={Polarized message-passing in graph neural networks},
  author={He, Tian Tian and Liu, Yang and Ong, Yew-Soon and others},
  journal={Artificial Intelligence},
  volume={331},
  pages={104129},
  year={2024}
}

@inproceedings{25defferrard2016,
  title={Convolutional neural networks on graphs with fast localized spectral filtering},
  author={Defferrard, Micha{\"e}l and Bresson, Xavier and Vandergheynst, Pierre},
  booktitle={Proceedings of the Advances in Neural Information Processing Systems},
  pages={3837--3845},
  year={2016}
}

@article{26zhang2024beyond,
  title={Beyond low-pass filtering on large-scale graphs via adaptive filtering graph neural networks},
  author={Zhang, Qi and Li, Jing Hua and Sun, Yan Feng and others},
  journal={Neural Networks},
  volume={169},
  pages={1--10},
  year={2024}
}

@inproceedings{27deb2024sea,
  title={{SEA-GWNN: simple and effective adaptive graph wavelet neural network}},
  author={Deb, Swakshar and Rahman, Sejuti and Rahman, Shafin},
  booktitle={Proceedings of the AAAI Conference on Artificial Intelligence},
  pages={11740--11748},
  year={2024}
}

@article{28ZhengZLLZ24NFGNN,
  title={Node-oriented spectral filtering for graph neural networks},
  author={Zheng, Shuai and Zhu, Zhen Feng and Liu, Zhi Zhe and others},
  journal={IEEE Transactions on Pattern Analysis and Machine Intelligence},
  volume={46},
  number={1},
  pages={388--402},
  year={2023}
}

@inproceedings{29zhang2024dig,
  title={{DiG-In-GNN: discriminative feature guided GNN-based fraud detector against inconsistencies in multi-relation fraud graph}},
  author={Zhang, Jing Hui and Xu, Zheng Jia and Lv, Dingyang and others},
  booktitle={Proceedings of the AAAI Conference on Artificial Intelligence},
  pages={9323--9331},
  year={2024}
}

@article{30huang2025correlation,
  title={Correlation information enhanced graph anomaly detection via hypergraph transformation},
  author={Huang, Chang Qin and Gao, Cheng Ling and Li, Ming and others},
  journal={IEEE Transactions on Cybernetics},
  year={2025},
  volume={55},
  number={6},
  pages={2865--2878}
}

@article{31lewicki2003approximation,
  title={Approximation by superpositions of a sigmoidal function},
  author={Lewicki, Grzegorz and Marino, Giuseppe},
  journal={Zeitschrift f{\"u}r Analysis und ihre Anwendungen},
  volume={22},
  number={2},
  pages={463--470},
  year={2003}
}

@inproceedings{32mcauley2013amazon,
  title={From amateurs to connoisseurs: modeling the evolution of user expertise through online reviews},
  author={McAuley, Julian John and Leskovec, Jure},
  booktitle={Proceedings of the International Conference on World Wide Web},
  pages={897--908},
  year={2013}
}

@inproceedings{33rayana2015yelpchi,
  title={{Collective opinion spam detection: Bridging review networks and metadata}},
  author={Rayana, Shebuti and Akoglu, Leman},
  booktitle={Proceedings of the ACM Sigkdd International Conference on Knowledge Discovery and Data Mining},
  pages={985--994},
  year={2015}
}

@article{34Elliptic,
  author       = {Mark Weber and
                  Giacomo Domeniconi and
                  Jie Chen and others},
  title        = {{Anti-Money Laundering in Bitcoin: Experimenting with Graph Convolutional
                  Networks for Financial Forensics}},
  journal      = {arXiv:1908.02591},
  year         = {2019}
}

@inproceedings{35tang2023gadbench,
 author = {Tang, Jian Heng and Hua, Feng Rui and Gao, Zi Qi and others},
 booktitle = {Proceeding of the Advances in Neural Information Processing Systems},
 pages = {29628--29653},
 title = {{GADBench: Revisiting and Benchmarking Supervised Graph Anomaly Detection}},
 year = {2023}
}

@inproceedings{36LiQLYL2019,
  author       = {Ao Li and
                  Zhou Qin and
                  Run Shi Liu and others},
  title        = {{Spam Review Detection with Graph Convolutional Networks}},
  booktitle    = {Proceedings of the {ACM} International Conference on Information and Knowledge Management},
  pages        = {2703--2711},
  year         = {2019},
}

@inproceedings{37liu2021,
  title={{Pick and choose: a GNN-based imbalanced learning approach for fraud detection}},
  author={Liu, Yang and Ao, Xiang and Qin, Zi Di and others},
  booktitle={Proceedings of the Web Conference},
  pages={3168--3177},
  year={2021}
}

@inproceedings{38chai2022,
  title     = {{Can Abnormality be Detected by Graph Neural Networks?}},
  author    = {Chai, Zi Wei and You, Si Qi and Yang, Yang and others},
  booktitle = {Proceedings of the International Joint Conference on Artificial Intelligence},
  pages     = {1945--1951},
  year      = {2022}
}

@inproceedings{39akiba2019optuna,
  title={{Optuna: A next-generation hyperparameter optimization framework}},
  author={Akiba, Takuya and Sano, Shotaro and Yanase, Toshihiko and others},
  booktitle={Proceedings of the ACM SIGKDD International Conference on Knowledge Discovery \& Data Mining},
  pages={2623--2631},
  year={2019}
}

@article{40maaten2008visualizing,
  title={{Visualizing data using t-SNE}},
  author={Maaten, Laurens van der and Hinton, Geoffrey},
  journal={Journal of Machine Learning Research},
  volume={9},
  pages={2579--2605},
  year={2008}
}

@inproceedings{41gao2023addressing,
author = {Gao, Yuan and Wang, Xiang and He, Xiang Nan and others},
title = {{Addressing Heterophily in Graph Anomaly Detection: A Perspective of Graph Spectrum}},
year = {2023},
booktitle = {Proceedings of the ACM Web Conference},
pages = {1528--1538}}

@article{45yuan2025comprehensive,
  title={{A comprehensive survey on GNN-based anomaly detection: taxonomy, methods, and the role of large language models}},
  author={Yuan, Zi Qi and Sun, Qing Yun and Zhou, Hao Yi and others},
  journal={International Journal of Machine Learning and Cybernetics},
  volume={16},
  number={7},
  pages={4407--4432},
  year={2025}
}

@inproceedings{49li2017radar,
  title={{Radar: Residual analysis for anomaly detection in attributed networks}},
  author={Li, Jun Dong and Dani, Harsh and Hu, Xia and others},
  booktitle = {Proceedings of the International Joint Conference on Artificial Intelligence},
  pages={2152--2158},
  year={2017}
}

@inproceedings{liu2026modeling,
title={{Modeling Spectral Energy Shifts in Spatio-Temporal Graph Anomaly Detection}},
author={Yi Lin Liu and Hong Chao Zhang and Ahmad Taha and others},
booktitle={Forty-third International Conference on Machine Learning},
year={2026}
}

@article{50ma2021comprehensive,
  title={A comprehensive survey on graph anomaly detection with deep learning},
  author={Ma, Xiao Xiao and Wu, Jia and Xue, Shan and others},
  journal={IEEE Transactions on Knowledge and Data Engineering},
  volume={35},
  number={12},
  pages={12012--12038},
  year={2021}
}

@article{WANG2025114039,
title = {Heterophily learning and global–local dependencies enhanced multi-view representation learning for graph anomaly detection},
journal = {Knowledge-Based Systems},
volume = {326},
pages = {114039},
year = {2025},
author = {Xiang Wang and Hao Dou and Zhenyu Meng}
}

@article{HUANG2026104338,
title = {Multi-faceted consistency data augmentation for graph anomaly detection},
journal = {Information Processing \& Management},
volume = {63},
number = {1},
pages = {104338},
year = {2026},
author = {Tai Ran Huang and Yi Li Wang and Qiu Tong Li and others}
}

@article{11278786,
  author={Guang, Ming Jian and Zhang, Rui and Cheng, Da Wei and others},
  journal={IEEE Transactions on Pattern Analysis and Machine Intelligence},
  title={{Homophily Edge Augment Graph Neural Network for High-Class Homophily Variance Learning}},
  year={2026},
  volume={48},
  number={3},
  pages={3835--3851}
  }

% Biography
%\bio{}
% Here goes the biography details.
%\endbio

%\bio{pic1}
% Here goes the biography details.
%\endbio

\end{document}